# Robust Multicentre Detection and Classification of Colorectal Liver Metastases on CT: Application of Foundation Models

Shruti Atul Mali[1], Zohaib Salahuddin[1], Yumeng Zhang[1], Andre Aichert[2], Xian Zhong[3], Henry C. Woodruff[1,4], Maciej Bobowicz[5], Katrine Riklund[6], Juozas Kupčinskas[7], Lorenzo Faggioni[8], Roberto Francischello[8], Razvan L Miclea[9], Philippe Lambin[1,4], *on behalf of EUCanImage working group*

[1] *Department of Precision Medicine, GROW - Research Institute for Oncology and Reproduction, Maastricht University, 6220 MD Maastricht, The Netherlands*

[2] *Digital Technology and Innovation, Siemens Healthcare GmbH, Erlangen, Germany.*

[3] *Department of Medical Ultrasonics, Institute of Diagnostic and Interventional Ultrasound, The First Affiliated Hospital of Sun Yat-sen University, Guangzhou, China*

[4] *Department of Radiology and Nuclear Medicine, GROW - Research Institute for Oncology and Reproduction, Maastricht University, Medical Center+, 6229 HX Maastricht, The Netherlands*

[5] *Department of Oncological, Transplant and General Surgery, Medical University of Gdańsk, Poland*

[6] *Umeå Center for Functional Brain Imaging (UFBI), Umeå University, Umeå, Sweden; Department of Diagnostics and Intervention, Umeå university, Umeå, Sweden.*

[7] *Department of Gastroenterology, Lithuanian University of Health Sciences, Kaunas; Institute for Digestive Research, Lithuanian University of Health Sciences, Kaunas, Lithuania.*

[8] *Department of Translational Research, Academic Radiology, University of Pisa, Pisa, Italy.*

[9] *Department of Radiology and Nuclear Medicine, Maastricht University Medical Center+, Maastricht, The Netherlands.*

# Abstract

**Background:**
Colorectal liver metastases (CRLM) remain a major cause of cancer-related mortality. Early, reliable detection on CT imaging is essential for curative treatment planning, yet the diagnostic performance of AI models often declines across scanners and institutions, limiting clinical generalizability.

**Purpose:**
To develop and evaluate a multi-center foundation model (FM)-based AI pipeline for patient-level classification and lesion-level detection of CRLM on contrast-enhanced CT, incorporating uncertainty quantification and explainability for clinically reliable decision-making.

**Materials and Methods:**
CT data from the EuCanImage consortium (n = 2,437) and TCIA_CRLM (n = 197, all CRLM) were used. Several pretrained FMs were benchmarked via logistic regression on extracted deep features, identifying UMedPT as the optimal encoder. A fine-tuned UMedPT + MLP model was trained for classification, and a Fully Convolutional One-Stage Object Detection (FCOS)-based head localized lesions. Uncertainty quantification (UQ) stratified cases by prediction confidence, decision curve analysis (DCA) evaluated clinical benefit, and Gradient-weighted Class Activation Mapping (Grad-CAM) provided insights towards interpretability.

**Results:**
The best classification model achieved AUC = 0.90 and sensitivity = 0.82 on the combined test set; sensitivity on the external TCIA cohort was 0.85. Excluding the most uncertain 20% of cases improved AUC to 0.91 and balanced accuracy to 0.86. DCA indicated superior net benefit over "treat-all" and "treat-none" strategies for threshold probabilities between 0.30–0.40. The lesion-level detector identified 69.1% of lesions overall, increasing from 30% (Q1) to 98% (Q4) across lesion-size quartiles. Grad-CAM maps showed strong correspondence between attention regions and metastases in high-confidence predictions. All source code is openly available.

**Conclusion:**
Foundation model-based pipelines enable robust, generalizable, and interpretable Colorectal liver metastases detection and classification on multiphasic CT, supporting the development of trustworthy AI tools for clinical oncology imaging.

**Keywords**: colorectal liver metastases, foundation model, artificial intelligence, computed tomography, multi-centre study, lesion detection, classification, uncertainty quantification

# 1. Introduction

Colorectal cancer (CRC) remains a major global health burden, ranking as the second most common type of cancer in women and third in men, with an estimated 3.2 million new cases worldwide annually, by the year 2040 [1]. Owing to the challenges of early detection of CRC, about 50 percent of the patients will ultimately progress to colorectal liver metastases (CRLM), which is one of the most aggressive types of liver tumors [2]. Precise diagnosis and localization of CRLM is important to inform treatment choices that can include surgical excision, systemic therapy, or locoregional treatment and directly affect patient outcomes [3,4]. Computed tomography (CT) is frequently used in the diagnosis and monitoring of CRLM [5,6]. Although the CT allows distinguishing between lesions and healthy liver tissue due to their contrast difference, it is challenging to differentiate between benign lesions (e.g, cysts) and malignant lesions since they may share similar imaging characteristics [7,8]. This leads to a time-consuming diagnostic procedure and a high likelihood of missing lesions. This has led to growing interest in the development of computer-aided diagnostic (CAD) tools that can aid radiologists in the classification and detection of focal liver lesions [9,10].

Radiomics has been given a lot of attention among other AI-based methods in medical image analysis [11,12]. Radiomics extracts hand-crafted quantitative descriptors from medical images to characterize tissue morphology, texture, and intensity, and has been explored for prognosis, treatment response assessment, and lesion classification in CRLM [13,14]. However, radiomic features are sensitive to scanner variability, reconstruction parameters, and acquisition protocols, requiring harmonization methods such as ComBat (corrects batch effects) for reproducibility in multi-centre studies [14,15]. ComBat is a statistical tool that corrects for scanner- or centre-related batch effects in features to improve reproducibility. The analysis of medical images with the help of deep learning has evolved after the discovery of AlexNet in natural image classification [16]. Since then, convolutional neural networks (CNNs) and architectures such as U-Net [17] have been used extensively in various applications such as classification, detection, and segmentation in medical imaging [18,19]. Deep learning has demonstrated great potential in automated lesion detection and characterization in CRLM [20,21]. However, deep learning algorithms typically need huge annotated datasets, and they do not tend to generalize to external settings and multi-centre cohorts, which restricts their applicability in practice [18,22].

Recently, Foundation models (FMs) have become a novel paradigm in medical image analysis [23]. FMs are trained on large-scale and heterogeneous datasets, usually employing self-supervised learning. They tend to produce generalizable feature representations that can be fine-tuned to achieve downstream tasks, especially with limited labelled data [24,25]. FMs have shown strong performance in oncology imaging in segmentation, classification, and biomarker discovery, and provide robust and more adaptable results than traditional supervised models [26]. Their ability to adapt to domain shifts and exploit broad prior knowledge makes them well-suited for multi-centre CRLM analysis.

Beyond accuracy, explainability and reliability are also crucial for clinical translation. Black box AI systems run the risk of compromising the clinician's trust and delaying integration into practice [27,28]. Integration of FMs with explainable AI techniques can not only offer robust performance, but also transparency, as a result of which a clinician can interpret the prediction and incorporate it into diagnostic decision-making [29,30]. Although FMs have been increasingly successful in other imaging applications, their use with CRLM has never been explored in a systematic manner. Moreover, the majority of previous research has focused on accuracy measures only without considering clinical interpretability, uncertainty quantification, and the decision-making potential. To be deployed in oncologic imaging, it is necessary not only to have high diagnostic accuracy but also to know the model uncertainty and to be able to visualize the anatomical regions driving predictions.

To fill these gaps, we developed a multi-center AI pipeline based on foundation models to perform patient-level classification and lesion-level detection of CRLM on contrast-enhanced CT. By using the EuCanImage consortium dataset, we benchmarked various pretrained FMs and fine-tuned UMedPT as the best encoder for downstream tasks. Using the best FM, we designed two complementary pipelines: (i) a patient-level classification model classifying CRLM from non-CRLM and negative cases, utilizing Gradient-weighted Class Activation Mapping (Grad-CAM) for interpretable visualization of model predictions; and (ii) a lesion-level detection model to locate metastatic liver lesions across multiple imaging cohorts. In addition, we also included uncertainty quantification (UQ), decision curve analysis (DCA), and explainability to evaluate the reliability of the model and the potential clinical benefit. This study is one of the most extensive analyses of FM-driven AI in CRLM, filling the gap between technical performance and clinical relevance in the direction of trustworthy and interpretable AI decision support in oncologic imaging.

# 2. Methods

Figure 1 shows the schematic overview of the pipeline. The workflow consists of three main stages: data curation, model development, and evaluation. Multicentre CT data and associated clinical information were curated within the EuCanImage infrastructure. TotalSegmentator was used to automatically segment the liver region, which was followed by localizing the liver to obtain liver patches. The images and masks were then resampled to the median voxel spacing of 1 × 1 × 2.5 mm, cropped to a fixed bounding box of 256 × 256 × 96 voxels based on the 95th percentile liver dimensions, and windowed between -37 and 171 HU to enhance liver contrast. Then, intensity normalization was applied on a per-scan basis to equalize the intensity values to a standard range.

The foundation model benchmarking, classification, and detection pipelines, and uncertainty quantification and explainability evaluation were part of model development. Both classification and detection were evaluated quantitatively and qualitatively, and additional evaluations were done to assess uncertainty-aware performance, decision-curve analysis (DCA), and model interpretability using Grad-CAM.

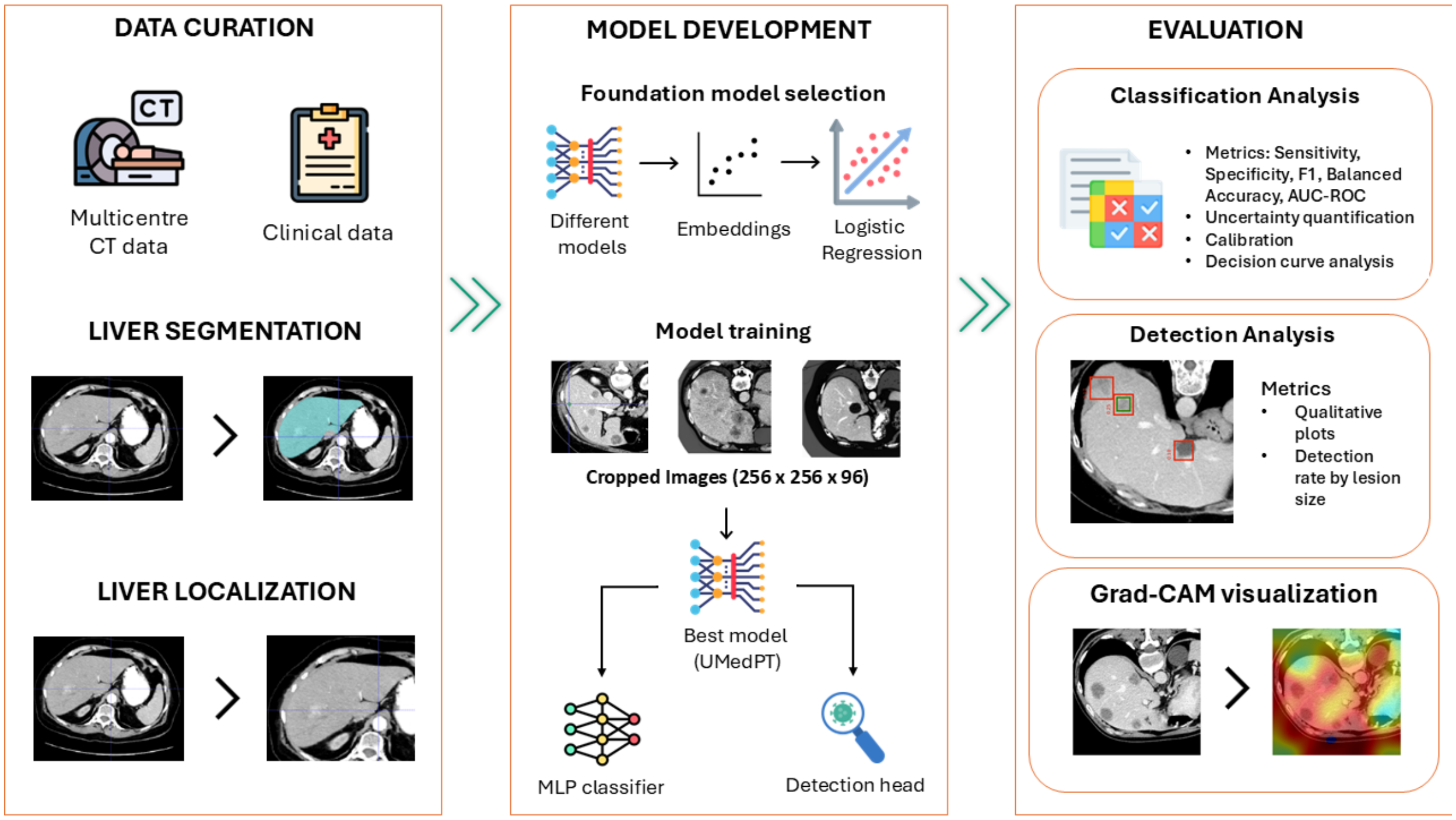

*Figure 1. Schematic overview of the proposed pipeline. The pipeline consisted of three stages: data curation (multicentre CT and clinical data), model development (foundation model selection and training classification and detection models), and evaluation (classification metrics, uncertainty quantification, decision curve analysis, lesion detection metrics, and Grad-CAM explainability). Preprocessing of images included resampling to 1 × 1 × 2.5 mm, cropping to 256 × 256 × 96 voxels, intensity windowing (-37 to 171 HU), and normalization per scan.*

## 2.1. Dataset:

This is a retrospective multi-site study that used the EuCanImage project data and publicly available Colorectal Liver Metastases (CRLM) data in the public database on The Cancer Imaging Archive (TCIA) [31]. EuCanImage is a European initiative that is focused on creating a cancer imaging platform to design and train AI algorithms on a variety of different tumors, such as colorectal cancer, lung cancer, and breast cancer. Imaging and clinical data were curated in the EuCanImage infrastructure that offers standard pipelines to process secure data and, in the meantime, adheres to the European regulations of data protection and privacy.

For the present work, we focused on the colorectal liver metastases (CRLM) use case, addressing the research question: “Can AI identify liver metastases in colorectal cancer from pre- and post-operative contrast-enhanced CT scans?” Patient inclusion criteria were: (i) Availability of contrast-enhanced CT imaging with liver coverage, (ii) Clinical and histopathological confirmation of diagnosis, and (iii) Diagnosis falling into one of three categories: CRLM, non-CRLM liver lesions (e.g., benign cysts, haemangiomas, or non-colorectal malignancies), or normal liver scans (negative). No specific exclusion criteria were applied. The final EuCanImage cohort consisted of 2,437 patients. A total of 2,437 patients with available CT imaging were included from four European hospitals (Kaunos, GUMED, Umeå, and Unipi). Each patient could contribute one or more CT scans, acquired either pre- or post-operatively as part of routine clinical workflows. Patients were stratified into three categories based on histopathological confirmation and radiology reports: (i) CRLM: colorectal liver metastases, (ii) non-CRLM: other liver lesions (e.g., benign cysts, non-colorectal malignancies), and (iii) negative: no liver lesions. For CRLM cases, up to three liver lesions per patient were annotated by expert radiologists. Since not all lesions were annotated, lesion masks were used only for lesion-level detection tasks, while patient-level binary labels (CRLM vs. non-CRLM/negative) were used for classification.

To improve robustness and evaluate generalizability, we additionally included the TCIA_CRLM collection (n ≈ 197 patients, portal venous phase CT) as an independent external dataset. This dataset consists only of CRLM-positive cases with full annotated lesions segmented semi-automatically. Therefore, it makes the data suitable for only lesion-level detection but not for balanced classification analyses.

The dataset was sampled on a per-patient, per-center basis to prevent scan-level leakage, and stratified sampling was done to maintain class balance. The final dataset splits were:

- Training/validation cohort (EuCanImage): 70% (n = 1,704 patients, 3,900 scans)
- Independent hold-out test cohort (EuCanImage): 30% (n = 733 patients, 1,692 scans)
- External test cohort (TCIA_CRLM): 197 patients (portal venous phase CT, all CRLM-positive).

The combined dataset of EuCanImage and TCIA_CRLM allowed evaluation of both patient-level classification (CRLM vs. non-CRLM/negative) and lesion-level detection tasks in a multi-institutional setting. Figure 1 shows the schematic of the study pipeline, including rigorous benchmarking of the FM-based AI methods and classification and detection tasks for CRLM.

### 2.1.1. Data preprocessing:

Standardized preprocessing was applied to all CT volumes to provide cross-site consistency. First, liver segmentation masks were automatically obtained with the help of TotalSegmentator [32], which provided liver volume delineations for all scans. All CT images and masks were resampled to the median voxel spacing of 1 x 1 x 2.5 mm. After resampling, each volume was cropped to a fixed bounding box set by the 95th percentile liver dimensions in the cohort, resulting in a standardized input size of 256 × 256 × 96 voxels. This cropping strategy made sure that the entire liver volume was always captured and removed irrelevant background. The intensities of voxels were windowed between -37 and 171 HU, which corresponds to the 0.5th-99.5th percentile in the EuCanImage contrast-enhanced CT training set to make liver parenchyma and lesions visible. This percentile-based windowing was estimated from all liver voxels across sites and contrast phases to minimize the effects of outliers and even out the intensity distributions for the liver ROI. The model was trained on all available contrast-enhanced CT phases (arterial, portal venous, and delayed) to provide complementary information for lesion characterization. Intensity normalization was applied on a per-scan basis to harmonize voxel intensity distributions across the dataset.

### 2.1.2. Foundation Model Benchmarking:

To identify the most suitable pretrained representation for CRLM classification, we systematically benchmarked five models that differed in pretraining strategy, dataset scale, and architecture. The evaluated models included CT-FM, FMCIB, Med3D, ModelsGenesis, and UMedPT (Table 1).

CT-FM [33] was trained using a modified SimCLR framework for self-supervised contrastive learning on approximately 148,000 CT scans, where intra-sample contrast was enforced by sampling patches within the same scan to capture spatial-semantic relationships; augmentations included random cropping, histogram shifting, and intensity scaling. Pretraining was performed for 500 epochs (best checkpoint obtained at epoch 449). The model uses a SegResEncoder backbone, a 3D convolutional residual encoder with roughly 77 million parameters and is intended as a general-purpose CT foundation model for segmentation, classification, and retrieval. Segmentation tasks were optimized with Dice and cross-entropy losses, whereas classification used binary cross-entropy with windowing tailored to CT-specific contrasts (e.g., blood, bone).

FMCIB (Foundation Model for Cancer Imaging Biomarkers) [23] was trained on ≈11,467 radiographic lesions from ~2,312 patients, spanning multiple cancer types. It employed a self-supervised contrastive learning objective, maximizing similarity between different augmentations of the same lesion and minimizing similarity across lesions. Its backbone is a 3D ResNet-50 adapted for volumetric lesion patches. The resulting embeddings are lesion-centric and have been shown to outperform prior baselines (e.g., Med3D, ModelsGenesis) for lesion classification and prognostic prediction, particularly in limited-data settings.

Med3D [34] was pretrained on the 3DSeg-8 dataset, comprising eight publicly available 3D segmentation datasets covering multiple organs and pathologies. It follows a supervised pretraining strategy. The family of encoders (3D ResNet-10/18/34) was trained for volumetric inputs and provides a strong initialization for downstream tasks, typically accelerating convergence and improving performance in both segmentation and classification (e.g., LiTS liver segmentation, lung nodule detection).

ModelsGenesis [35] introduced self-supervised learning on 3D medical volumes, through image restoration and transformation tasks (e.g., recovering masked patches, outpainting, inpainting, and context restoration). Pretraining was performed on large-scale, unlabelled CT and MRI volumes. The network adopts a 3D U-Net-like encoder-decoder (with residual connections in some implementations), yielding a robust initialization that consistently outperforms training from scratch for segmentation and classification when annotations are limited.

UMedPT [36] represents a general-purpose medical foundation model trained in a multi-task supervised framework across 15 large-scale datasets, covering classification, segmentation, detection, and survival prediction and spanning multiple modalities (CT, MRI, PET, X-ray). It employs a hierarchical Swin Transformer encoder with shifted windows coupled to task-specific heads (e.g., segmentation decoders, classification layers). Pretraining at this scale (≈4 million images across the constituent datasets) supports broad transferability for both 2D and 3D medical imaging tasks.

***Table 1:*** *Summary table of foundation models.*

| Model | Model type | Architecture | Pretraining strategy | Pretraining dataset |
|---|---|---|---|---|
| CT-FM | Foundation | SegResEncoder (3D convolutional residual encoder) | Self-supervised (modified SimCLR, intra-sample contrastive) | ≈148,000 CT scans |
| FMCIB | Foundation | 3D ResNet-50 backbone | Self-supervised (contrastive lesion-centered SSL) | ≈11,467 CT lesions from ~2,312 patients |
| Med3D | Supervised | 3D ResNet family (ResNet10/18/34) | Supervised segmentation | 3DSeg-8 (8 public segmentation datasets, multi-organ CT/MRI) |
| Models Genesis | Self-supervised | 3D U-Net-like encoder-decoder | Self-supervised (image restoration, inpainting, outpainting, context prediction) | Large-scale unlabelled CT and MRI volumes (incl. 623 chest CTs) |
| UMedPT | Foundation | Swin Transformer encoder + task-specific heads | Multi-task supervised pretraining across classification, segmentation, detection, and survival prediction | ≈4 million images across 15 biomedical datasets (CT, MRI, PET, X-ray) |

For each model, deep feature embeddings were extracted from the pre-processed CT volume. To address class imbalance, we employed the Synthetic Minority Oversampling Technique (SMOTE) [37] to generate balanced training folds. Before classification, features were reduced in dimensionality using principal component analysis (PCA), with the number of retained components optimized during model selection. Then, a logistic regression classifier was built to carry out a binary classification task(CRLM vs. non-CRLM/negative).

Hyperparameter optimization was done using Optuna [38] with a search space that consisted of the number of components for PCA, regularization parameter (C), type of penalty (L1, L2, elastic net), solver, class weights (balanced), maximum iterations (varied to ensure solver convergence), and mixing ratio of elastic net. Each candidate configuration was evaluated using 10-fold stratified cross-validation on the training set, and the mean validation AUC served as the optimization objective.

The best-performing configuration was retrained on the entire training dataset and evaluated on the independent test sets (combined and subgroup analyses). The foundation model achieving the highest validation AUC was selected for downstream development of the classification and detection models

## 2.2. Model Development:

### 2.2.1. Classification (patient-level):

For the classification task, the selected FM backbone (UMedPT) was used as the feature encoder. The network was fine-tuned by unfreezing a few encoder blocks, and an MLP classification head was attached to the extracted features. The MLP accepted a 512-dimensional embedding produced by the UMedPT grouper module and consisted of two fully connected layers (512 → 256 and 256 → 128), each followed by instance normalization, a LeakyReLU activation (negative slope = 0.01), and dropout (p = 0.2) for regularization. The final fully connected layer mapped the 128-dimensional representation to a single output logit, corresponding to the binary classification task (CRLM vs. non-CRLM/negative). The model was trained using 5-fold cross-validation on the EuCanImage training set, with patient-level stratification to preserve class balance. Predictions for the independent test sets were obtained by averaging probabilities across the 5 trained folds (ensemble prediction). The inference was done at the patient level. This design provided sufficient non-linear capacity to discriminate between patient classes while incorporating regularization to reduce overfitting. The classification head was trained jointly with the UMedPT encoder (Figure 2).

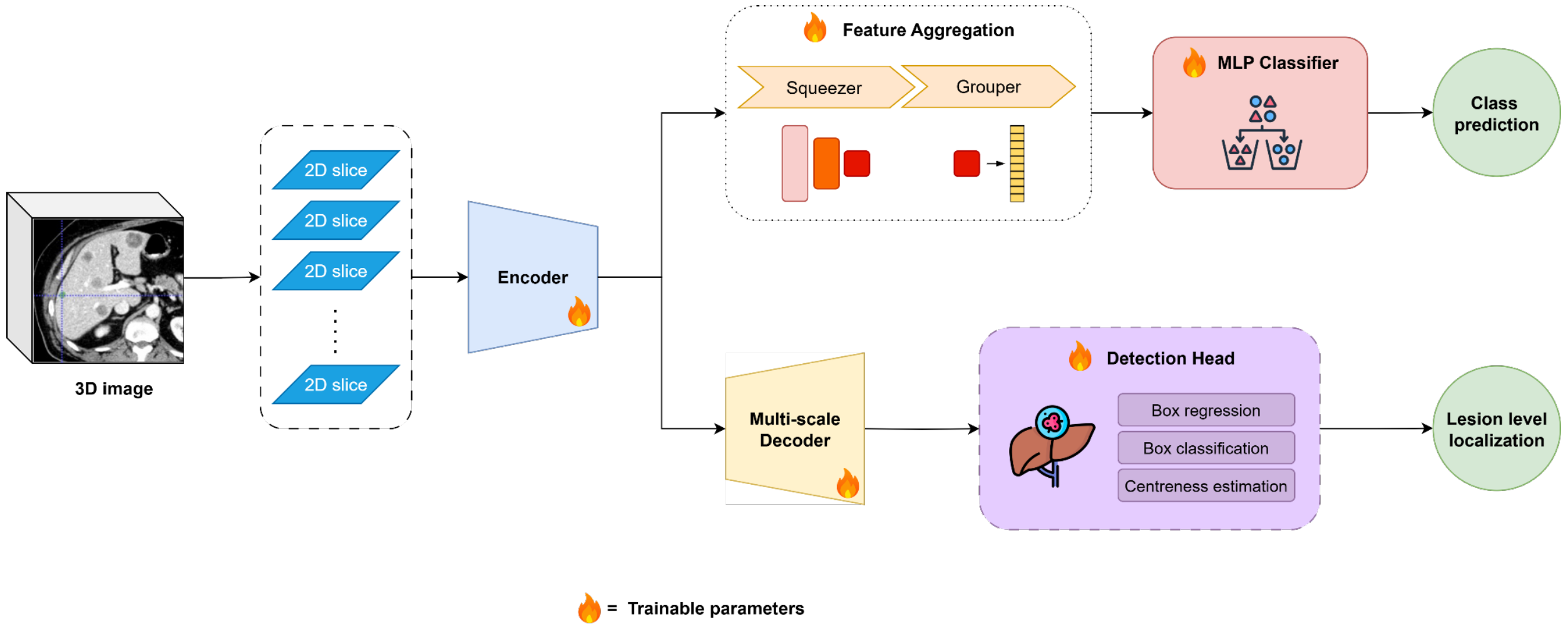

*Figure 2. Overview of the proposed UMedPT-based CRLM classification and detection framework. The model integrates two complementary pipelines: (1) a classification branch that aggregates deep features extracted from the pretrained UMedPT encoder and predicts the patient-level probability of CRLM via an MLP classifier, and (2) a lesion-level detection branch using a multi-scale decoder and FCOS-style detection head for bounding-box regression, classification, and centerness estimation.*

### 2.2.2. Uncertainty Quantification and Calibration:

Uncertainty quantification (UQ) was performed to evaluate the reliability of model predictions, following the approach of Alves et al. [39]. Prediction uncertainty was estimated by calculating the variance across five independently trained UMedPT+MLP folds. The ensemble-averaged probability was considered as the final prediction, and the variance across all 5 folds was used as an uncertainty score per patient (lower variance indicates higher confidence). This uncertainty quantification was used to rank the patients and stratify cohorts into a certain group (CG) and an uncertain group (UG) at different percentile thresholds (10%-90%). At each threshold, performance metrics such as area under the ROC curve (AUC) and balanced accuracy were recalculated separately within the CG and UG subsets. Ninety-five percent confidence intervals were estimated using 1,000 bootstrap resamples at the patient level.

Calibration was assessed on the combined test set using calibration curves and metrics, including Brier score, calibration slope, and intercept. Perfect calibration corresponds to slope = 1 and intercept = 0; deviations indicate over- or under-confidence.

### 2.2.3. Decision Curve Analysis:

To assess potential clinical utility, decision curve analysis (DCA) was performed using the ensemble-averaged predictions from the 5-fold UMedPT+MLP model. Net benefit was calculated as:

$$Net\ benefit\ = \frac{TP}{n} - \frac{FP}{n} \times \frac{p}{1-p}$$

Where TP and FP are the number of true and false positives, n is the total number of patients, and p is the threshold probability. Curves were generated across the full range of threshold probabilities and compared against “treat-all” and “treat-none” strategies.

### 2.2.4. Detection (lesion-level):

For CRLM lesion detection, the UMedPT encoder was extended with a fully convolutional one-stage (FCOS) detection head adapted for volumetric CT data, but in a 2D slice-wise manner (Refer to Figure 2). A feature pyramid network (FCOS decoder) generated multi-scale representations from the encoder outputs, and these were passed to the detection head (MTLFCOSHead). The detection head consisted of three parallel convolutional branches for classification, bounding box regression, and centerness prediction. Each branch was composed of two stacked 3×3 convolutional layers with normalization and non-linear activations. The classification branch predicted lesion probability maps, the regression branch estimated distances to lesion boundaries for bounding box construction, and the centerness branch penalized off-centre predictions to improve localization quality. The model was optimized with a multi-task objective comprising focal loss for classification ($\gamma = 2.0$, $\alpha = 0.25$, weighted by 1/log(2)), IoU loss for bounding box regression (weight = 1.0), and binary cross-entropy for centerness prediction (weight = 0.2). This anchor-free method was able to detect lesions of different sizes effectively. The model was weakly supervised using partially annotated lesions (up to three lesions per CRLM patient). The model was assessed at a lesion level. A lesion was counted as detected when the IoU was more than 0.025. Bounding boxes predicted with an IoU $\leq$ 0.025, relative to all ground-truth lesions, were considered false positives, and false positive rates (FPR) were reported per image and per ground-truth lesion. In order to explore the impact of the lesion size, lesions were classified into quartiles based on their volume. The volumes were calculated by multiplying the number of voxels and the size of the voxels (1 × 1 × 2.5 mm, ≈0.0025 $cm^3$ per voxel). The distribution of the lesion volumes in the test set was used to define the boundaries of the quartiles, and the rate of detection by each of the quartiles (Q1-Q4) was reported separately. This stratification enabled the determination of the sensitivity of lesion size in the TCIA CRLM cohort, since it was a fully annotated dataset.

### 2.2.5. Explainability:

In order to achieve interpretability, we utilized Gradient-weighted Class Activation Mapping (Grad-CAM) for the classification pipeline. Grad-CAM is an explainability method that generates visual explanations of convolutional neural networks, based on the gradient of target class scores relative to the final convolutional feature maps of the network. This results in heatmaps that highlight the regions of the input most influential in the model's decision-making process [40]. In our implementation, Grad-CAM was applied to the UMedPT encoder during inference for classification. Heatmaps were generated from the 'norm1' layer of the first SwinTransformerBlockV2 in the seventh encoder block. To provide robust and stable interpretability, we generated Grad-CAM maps for each of the five cross-validation folds and then computed a consensus heatmap by averaging the fold-level heatmaps. These consensus maps were overlaid onto the original CT slices, enabling visualization of the anatomical regions that contributed most strongly to predictions of CRLM versus non-CRLM/negative.

### 2.2.6. Training:

All models were implemented in PyTorch with MONAI for data loading and augmentation and trained on NVIDIA H100 GPUs using mixed precision (AMP) to accelerate training and reduce memory consumption.

Training workflows for both the classification and detection pipelines are summarized in Figure 2. For the classification model, training followed a 5-fold cross-validation design with patient-level stratification. Each fold was optimized with focal binary cross entropy loss ($\gamma = 3.0$, $\alpha = 0.7$) to address class imbalance. The optimizer was AdamW with a learning rate of 5e-4. Training was run for a maximum of 1000 epochs, with early stopping if no improvement in validation loss was observed for 10 epochs. The batch size was set to 12. During training, performance was monitored using accuracy, AUC, precision, recall, and F1-score, and the best checkpoint was saved based on validation loss. Final test set predictions were obtained by averaging probabilities across the 5 folds (ensemble prediction).

For the detection model, training was based on the FCOS anchor-free detection framework. The detection head was optimized with a multi-task loss comprising focal loss for classification ($\gamma = 2.0$, $\alpha = 0.25$, weighted by 1/log(2)), IoU loss for bounding box regression (weight = 1.0), and binary cross-entropy for centeredness prediction (weight = 0.2). The optimizer was AdamW with a learning rate of 1e-4, trained for up to 1000 epochs with early stopping (patience = 10). The batch size was 12, reflecting the higher memory demand for the detection module.

### 2.2.7. Loss functions:

For classification, the primary objective was focal loss [41] to address class imbalance. Unlike BCE, focal loss introduces a modulating factor that reduces the relative loss for well-classified examples and focuses training on hard or misclassified cases. The loss is defined as:

$$L_{Focal} = -\alpha_t (1 - p_t)^{\gamma} \ log(p_t)$$

where $p = \sigma(\hat{y})$ . Here, $\alpha \in [0,1]$ balances positive and negative examples while $\gamma \geq 0$ adjusting the focus on misclassified cases.

For detection, the training objective followed the anchor-free FCOS framework, combining three components:

- Classification loss: focal loss, encouraging correct detection of lesion voxels while handling severe class imbalance between lesion and background.

- Bounding box regression loss: IoU loss [42], which directly optimizes the overlap between predicted and ground-truth bounding boxes. It is defined as:

$$L_{IoU} = 1 - \frac{| B_p \cap B_{gt} |}{| B_p \cup B_{gt} |}$$

  Where $B_p$ and $B_{gt}$ are predicted and ground-truth bounding boxes.

- Centreness loss: Binary cross-entropy loss, encouraging predictions close to lesion centres.

$$L_{BCE} = -[\, y_i \ log \, \sigma(\hat{y}_i) + (1 - y_i) \ log \, (1 - \sigma(\hat{y}_i)) \,]$$

  Where $y_i \in \{0, 1\}$ is the ground-truth label, $\hat{y}_i$ the predicted logit, and $\sigma$ the sigmoid activation.

## 2.3. Evaluation metrics:

For classification, metrics like balanced accuracy, area under the receiver operating characteristic curve (AUC), sensitivity, specificity, precision, and F1-score were reported.

- **Sensitivity (Recall):** the proportion of CRLM patients correctly identified,

$$Sensitivity = \frac{TP}{TP + FN}$$

  Where $TP \ and \ FN$ are true positives, and false negatives.

- **Specificity:** the proportion of non-CRLM/negative patients correctly identified,

$$Specificity = \frac{TN}{TN + FP}$$

Where $TN \text{ and } FP$ are true negatives and false positives.

- **Balanced accuracy**: the average of sensitivity and specificity, which accounts for class imbalance,

$$Balanced\ Accuracy = \frac{Sensitivity + Specificity}{2}$$

- **Area Under the Receiver Operating Characteristic Curve (ROC AUC):** The ROC AUC measures the model's ability to discriminate between CRLM and non-CRLM/negative patients and across all possible classification thresholds. It is calculated as the area under the ROC curve by plotting the true positive rate (sensitivity) against the false positive rate (1 − specificity).

- **Precision (Positive Predictive Value)**: the proportion of patients with predicted CRLM that were correctly classified,

$$Precision = \frac{TP}{TP + FP}$$

- **F1-score**: the harmonic mean of precision and recall,

$$F1 = 2 * \frac{Precision * Sensitivity}{Precision + Sensitivity}$$

For lesion detection, performance was assessed using appropriate metrics for bounding box regression:

- **Detection rate**: the rate that defines the fraction of annotated lesions that were detected with IoU above a given threshold,

$$Detection\ Rate = \frac{\#Lesions\ detected\ at\ IoU \geq threshold}{\#Total\ ground\ truth\ lesions}$$

- **False positive rate**: The number of predicted bounding boxes that do not overlap with any ground-truth lesions (IoU ≤ 0.025), expressed either per image or per lesion,

$$FPR_{image} = \frac{\#Total\ predicted\ lesions\ -\ \#Total\ ground\ truth\ lesions}{\#Total\ images}$$

$$FPR_{lesion} = \frac{\#Total\ predicted\ lesions\ -\ \#Total\ ground\ truth\ lesions}{\#Total\ ground\ truth\ lesions}$$

- **Size-stratified detection performance**: lesions were grouped by quartiles of lesion volume (in voxels), and detection rates were computed within each quartile (Q1, Q2, Q3, Q4) to evaluate sensitivity to lesion size.

## 2.4. RQS 2.0 Assessment:

The methodological quality of the proposed pipeline was assessed with the help of the Radiomics Quality Score (RQS 2.0) framework [43]. All of the RQS requirements, including data preparation, model development, validation, and trustworthiness, were evaluated using the official scoring requirements. The cumulative score was mapped to the corresponding Radiomics Readiness Level (RRL) to quantify methodological maturity. Detailed scoring criteria and evidence mapping are provided in Supplementary Table 2.

# 3. Results

## 3.1. Clinical and Imaging data:

The final EuCanImage cohort included 2,437 patients across four centres: KAUNOS (n = 490), GUMED (n = 848), UMEA (n = 401), and UNIPI (n = 698), plus an additional 197 patients from the external TCIA_CRLM dataset (Table 2). The mean patient age ranged between 63.5 ± 13.3 and 68.2 ± 12.5 years, with a balanced gender distribution (female 35-50%; male 50-61%). The TCIA_CRLM cohort had a mean age of 59.7 ± 12.3 years and included 117 females (59.4%) and 80 males (40.6%). Among EuCanImage patients, 710 (29%) were diagnosed with colorectal liver metastases (CRLM), 1,004 (41%) presented with non-CRLM lesions, and 723 (30%) were negative for healthy subjects. The TCIA_CRLM dataset consisted exclusively of CRLM-positive cases (n = 197, 100%). The total number of annotated CRLM lesion masks was 796 cases across both datasets: 599 from EuCanImage centres and 197 from TCIA_CRLM.

Table 2: Demographic and imaging overview of the EuCanImage colorectal liver metastasis (CRLM) cohort. Values are reported as mean ± standard deviation (SD) or number (percentage). The table summarizes patient distribution, age, gender, and counts of CRLM, non-CRLM, and negative cases, along with the total number of lesion masks available per site.

| | KAUNOS | GUMED | UMEA | UNIPI | TCIA_CRLM |
|---|---|---|---|---|---|
| **Clinical data** | | | | | |
| **Age:** Mean (SD) | 63.5 ± 13.3 | 66.3 ± 11.8 | 66.6 ± 10.1 | 68.2 ± 12.5 | 59.7 ± 12.3 |
| **Gender** | | | | | |
| Female | 242 (49.4%) | 363 (42.8%) | 142 (35.4%) | 348 (49.9%) | 117 (59.4%) |
| Male | 248 (50.6%) | 485 (57.2%) | 247 (61.6%) | 350 (50.1%) | 80 (40.6%) |
| unknown | - | - | 12 (3.0%) | - | - |
| **Imaging data** | | | | | |
| Total patients | 490 | 848 | 401 | 698 | 197 |
| CRLM cases | 237 (48.36%) | 353 (41.63%) | 27 (6.73%) | 93 (13.32%) | 197 (100%) |
| non-CRLM cases | 149 (30.41%) | 209 (24.64%) | 374 (93.26%) | 272 (38.96%) | 0 (0.0%) |
| Negative cases | 104 (21.22%) | 286 (33.72%) | 0 (0.0%) | 333 (47.70%) | 0 (0.0%) |
| Lesion masks (CRLM) | 216 (44.08%) | 308 (36.32%) | 26 (6.48%) | 49 (7.02%) | 197 (100%) |

## 3.2. Foundation model benchmarking:

To identify the most suitable pretrained backbone for CRLM classification, we compared five candidate foundation models using logistic regression applied to their extracted deep features (Table 3). In terms of AUC, values ranged from 0.66 (Med3D + LR) to 0.81 (UMedPT + LR). FMCIB + LR and ModelsGenesis + LR achieved intermediate values (0.70 and 0.73, respectively), while CT-FM + LR reached 0.74. Balanced accuracy showed the same pattern, with UMedPT + LR having the largest value (0.74) as opposed to 0.62-0.67 with other models.

In terms of sensitivity, which measures the capacity of the model to identify CRLM patients accurately, the overall performance in the LR benchmark was typically low, ranging between 0.51 and 0.62 (Med3D + LR at the lower and UMedPT + LR at the upper). The specificity was also uniformly higher among the models (at most 0.87 in UMedPT + LR), reflecting more discrimination of non-CRLM/negative cases. Precision ranged from 0.43 (Med3D + LR) to 0.65 (UMedPT + LR), and F1-scores from 0.47-0.63. Overall, the UMedPT model provided the best classification performance across all metrics.

These results indicate that UMedPT provided the most discriminative and robust feature space for CRLM classification across all classification metrics. In contrast, Med3D + LR and FMCIB + LR consistently underperformed, while CT-FM + LR and ModelsGenesis + LR achieved intermediate results.

Table 3: Benchmarking of logistic regression (LR) classifiers applied to deep feature embeddings extracted from five candidate foundation models. Reported metrics include area under the ROC curve (AUC), balanced accuracy, sensitivity, specificity, precision, and F1-score with 95% confidence intervals. UMedPT + LR achieved the highest overall performance across metrics.

| Model | AUC (CI) | Balanced Accuracy (CI) | Sensitivity (CI) | Specificity (CI) | Precision (CI) | F1-score (CI) |
|---|---|---|---|---|---|---|
| FMCIB + LR | 0.70 (0.66-0.74) | 0.63 (0.59-0.67) | 0.52 (0.45-0.58) | 0.74 (0.71-0.78) | 0.45 (0.4-0.52) | 0.48 (0.42-0.54) |
| UMedPT + LR | **0.81 (0.77-0.84)** | **0.74 (0.71-0.78)** | **0.62 (0.55-0.68)** | **0.87 (0.84-0.9)** | **0.65 (0.58-0.72)** | **0.63 (0.58-0.68)** |
| CT-FM + LR | 0.74(0.70-0.78) | 0.67 (0.63-0.71) | 0.58 (0.52-0.65) | 0.76 (0.72-0.8) | 0.5 (0.43-0.56) | 0.54 (0.48-0.59) |
| Models Genesis + LR | 0.73 (0.69-0.77) | 0.64 (0.60-0.68) | 0.55 (0.49-0.62) | 0.73 (0.69-0.76) | 0.45 (0.4-0.51) | 0.5 (0.44-0.55) |
| Med3D + LR | 0.66 (0.61-0.70) | 0.62 (0.58-0.66) | 0.51 (0.45-0.58) | 0.72 (0.68-0.76) | 0.43 (0.37-0.49) | 0.47 (0.41-0.52) |

To assess potential scanner-related bias, ComBat feature harmonization was applied across sites using the UMedPT features. However, no performance improvement was observed (AUC = 0.72 vs. 0.81 pre-harmonization), indicating that the foundation model already yielded site-robust representations (see Supplementary Material).

## 3.3. Classification performance:

Building on the benchmark, we developed a dedicated classification pipeline using UMedPT as the backbone with a fine-tuned MLP head. Although training used all available contrast phases to enhance robustness, evaluation was restricted to the PV phase, which provides the most consistent lesion-to-parenchyma contrast. In our dataset, PV-phase scans were available for 55% of training cases and 60% of cases for the combined test set (EuCanImage + TCIA-CRLM).

Using the optimal operating threshold (0.4369) derived from Youden's index, the model achieved an AUC of 0.90 (95% CI: 0.87-0.92) and a balanced accuracy of 0.83 (95% CI: 0.80-0.85) on the combined EuCanImage + TCIA-CRLM test cohort (n = 930) (Table 4). Sensitivity and specificity were well balanced (0.82 and 0.83, respectively), while precision (0.79) and F1-score (0.81) reflected robust discrimination between CRLM and non-CRLM/negative cases. Figure 3 illustrates the receiver operating characteristic (ROC) curves of the five cross-validation folds, and the ensemble test set, where the values of the area under the curve are consistently high in each fold and the generalization on the combined test set (EuCanImage test set + TCIA_CRLM). The associated confusion matrix of the portal venous phase test data (Figure 3) confirms the balanced distribution of true and false predictions, which is in accordance with the reported sensitivity and specificity values.

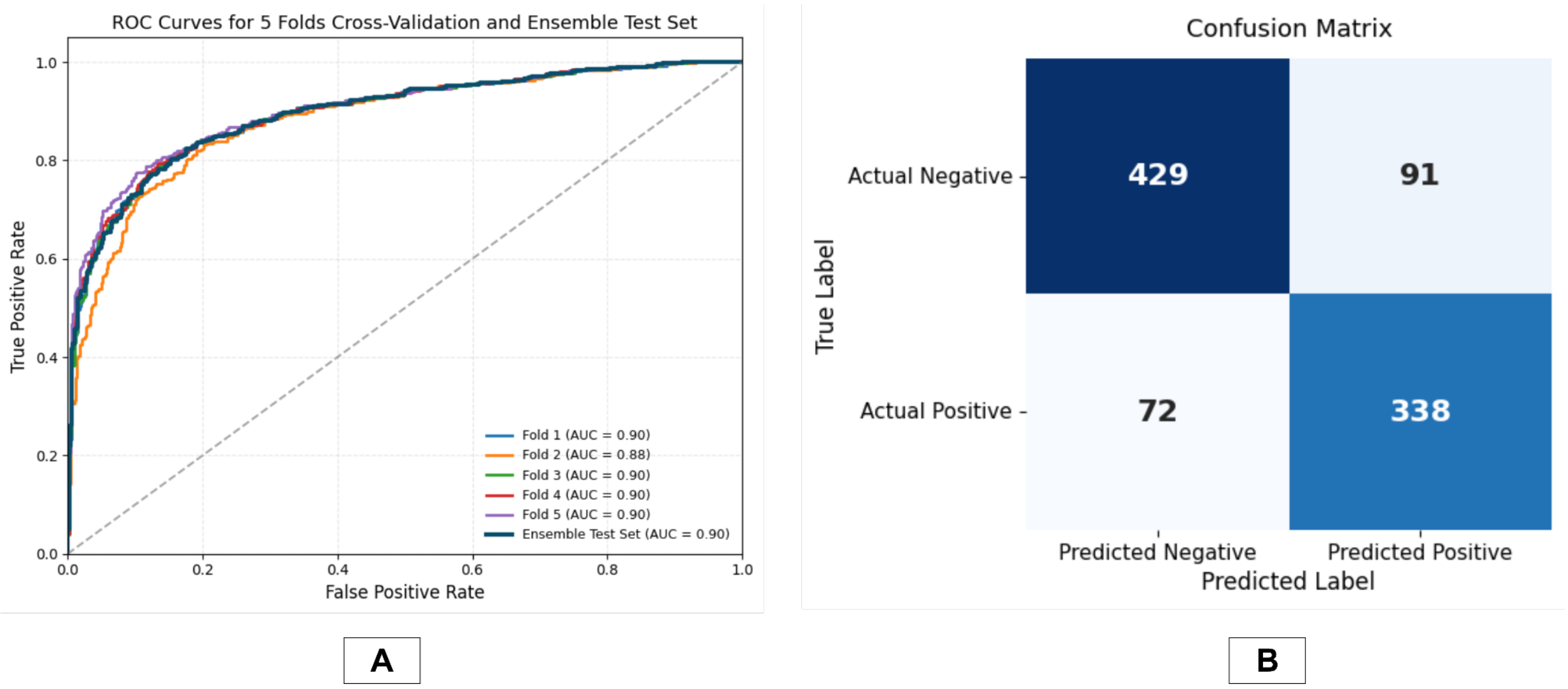

*Figure 3. Classification performance of UMedPT + MLP model on the combined test set (portal venous phase). (A) ROC curves of five cross-validation folds and the ensemble model, which depict consistently high AUC values across folds and robust performance on the combined test set. (B) Confusion matrix, at the selected operating threshold (0.4369), which presents the distribution of correctly and incorrectly classified cases (TN = 429, FP = 91, FN = 72, TP = 338).*

Subgroup analyses ensured similar performance across datasets. The model gave an AUC of 0.88 (95% CI: 0.85-0.91), balanced accuracy of 0.82 (95% CI: 0.78-0.85), sensitivity of 0.82, and specificity of 0.81 for the EuCanImage test set (n = 733). On the external TCIA_CRLM cohort (n = 197; all-positive cases), only the recall metric was applicable, where the model achieved a sensitivity of 0.85 (95% CI: 0.80-0.90). The MLP model was evaluated per-site, and it gave a stable performance across each of these centres, with AUC values ranging from 0.79 to 0.92 and balanced accuracy between 0.77 and 0.83. The model achieved the highest performance at KAUNOS (AUC = 0.92, sensitivity = 0.90) and GUMED (AUC = 0.88, balanced accuracy = 0.83), indicating consistent generalization within these cohorts. UMEA and UNIPI sites showed relatively lower F1-scores (0.37 and 0.52, respectively), reflecting the impact of severe class imbalance and smaller positive case counts (8/121 for UMEA and 28/210 for UNIPI). The model showed a good discrimination ability (AUC ≥ 0.79) and balanced accuracy was in the range 0.71-0.83, giving decent sensitivity-specificity trade-offs across multiple centres. The GUMED cohort showed the best performance amongst other sites, likely due to its majority representation in the training set, which may have reduced site-related feature shifts.

Table 4: Performance of the fine-tuned UMedPT + MLP classifier on the combined EuCanImage + TCIA_CRLM test cohort (n = 930) and in subgroup analyses on the EuCanImage test set (n = 733), EuCanImage sites (KAUNOS, GUMED, UMEA, UNIPI), and TCIA_CRLM subset (n = 197, all CRLM cases, portal venous phase only). Results are reported for portal venous phase inputs at the optimal threshold of 0.4369. Metrics include AUC, balanced accuracy, sensitivity, specificity, precision, and F1-score with 95% confidence intervals.

| Test set | AUC (CI) | Balanced Accuracy (CI) | Sensitivity (CI) | Specificity (CI) | Precision (CI) | F1-score (CI) |
|---|---|---|---|---|---|---|
| Test Set (n=930) | 0.9 (0.87-0.92) | 0.83 (0.80-0.85) | 0.82 (0.78-0.86) | 0.83 (0.79-0.86) | 0.79 (0.75-0.83) | 0.81 (0.78-0.84) |
| **Datasets** | | | | | | |
| EuCanImage (n=733) | 0.88 (0.85-0.91) | 0.82 (0.78-0.85) | 0.82 (0.76-0.87) | 0.81 (0.78-0.85) | 0.64 (0.59-0.7) | 0.72 (0.67-0.76) |
| TCIA_CRLM (n=197) | - | - | 0.85 (0.8-0.9) | - | - | - |
| **Centers** | | | | | | |
| KAUNOS (n=147) | 0.92 (0.87-0.96) | 0.79 (0.73-0.86) | 0.90 (0.83- 0.97) | 0.68 (0.57-0.79) | 0.73 (0.64-0.82) | 0.81 (0.73-0.87) |
| GUMED (n=255) | 0.88 (0.83-0.92) | 0.83 (0.78-0.88) | 0.81 (0.74-0.89) | 0.85 (0.79-0.91) | 0.79 (0.71-0.87) | 0.80 (0.74-0.86) |
| UMEA (n=121) | 0.87 (0.77-0.976 | 0.71 (0.53-0.89) | 0.50 (0.17-1.00) | 0.92 (0.86-0.96) | 0.29 (0.06-0.56) | 0.37 (0.1-0.61) |
| UNIPI (n=210) | 0.79 (0.68-0.9) | 0.77 (0.67-0.86) | 0.68 (0.50-0.84) | 0.86 (0.81-0.91) | 0.42 (0.29-0.57) | 0.52 (0.39-0.66) |

## 3.4. Misclassification Analysis:

To better understand model behavior, we examined misclassifications across CRLM, non-CRLM, and Negative subgroups for the combined test set; for summary reporting, non-CRLM and Negative were grouped as "Other." Most false positives originated from the non-CRLM group (60 out of 303 non-CRLM cases, 19.8%) and, to a lesser extent, the Negative group (31 out of 217 Negative cases, 14.3%), yielding an overall false-positive rate among "Other" of 17.5% (91 out of 520). In the CRLM group, 17.8% (73 of 410 CRLM cases) were missed (predicted as Other,) giving a sensitivity of 82.2%.

In order to study classification failures in more detail, we examined the 30 misclassified cases in the TCIA_CRLM test set, all of which represented false negative cases (i.e., CRLM-positive patients that were classified as Other). In all these patients, the total number of lesions annotated was 49, median of one lesion per case (range: 1-4). The mean lesion volume was 4.7 $cm^3$ (median: 1.1 $cm^3$, IQR: 0.3-3.5 $cm^3$). Almost half of all lesions (47%) were less than 1 $cm^3$, and one-third of the patients (10/30) had sub-centimeter lesions. A small subgroup (6/30) had at least one large lesion (>10 $cm^3$). These results showed that misdiagnosed TCIA_CRLM cases were often associated with small or subtle metastases, although a few cases also had larger, possibly poorly visible lesions.

A similar pattern could be observed in the EuCanImage portal-venous subset, in which 134 test cases were misclassified, out of which only 66 had available tumor annotations. In these patients, about 140 annotated lesions were found with a median of two lesions per case (range: 1-5). The mean lesion volume was 6.7 cm³ (median: 1.0 cm³, IQR: 0.3-3.1 cm³) and almost half of the total annotated lesions were less than 1 cm³. Larger lesions (≥10 cm³) represented approximately one-fifth of the total. Since not all lesions were annotated, quantitative sensitivity of detection could not be accurately estimated, but the distribution observed indicates that small or less conspicuous metastases were once again common among partially annotated misclassified cases. Probability estimates of the classification (UMedPT + MLP) model were tested with calibration curves on the combined model test set (Refer to Figure 1 in the Supplementary material). The model had moderate calibration of the joint test set (Brier = 0.17, slope = 3.4, intercept = 0.15), which showed a bit overconfident estimates of the probability.

## 3.5. Uncertainty quantification:

We stratified the subjects into a group of certainty (CG) and an uncertain group (UG) at different thresholds of certainty to determine the relationship between model confidence and predictive reliability (Figure 4). The curves indicate balanced accuracy and AUC with respect to the percentage of subjects retained in the CG. Expectedly, the performance was always better with high certainty, and the CG outperformed the UG on all measures. CG was maintaining comparable accuracy and AUC values as the overall model performance, whereas UG gradually decreased and became less stable. The shaded regions represent 95% bootstrap confidence intervals. At the threshold where 80 percent of the subjects were retained in the CG (i.e., the most uncertain 20 percent were excluded), the model had a balanced accuracy value of 0.86 (95 percent CI: 0.83-0.88) and AUC = 0.91 (95 percent CI: 0.89-0.93). It means that the accuracy and AUC were significantly increased by eliminating the 20 percent most uncertain cases. Conversely, the uncertain group (UG) of the same threshold exhibited a sharp reduction in performance (balanced accuracy = 0.51 (95% CI: 0.50-0.53), AUC = 0.74 (95% CI: 0.65-0.81)), which confirms that uncertainty quantification is effective to distinguish between reliable and unreliable predictions.

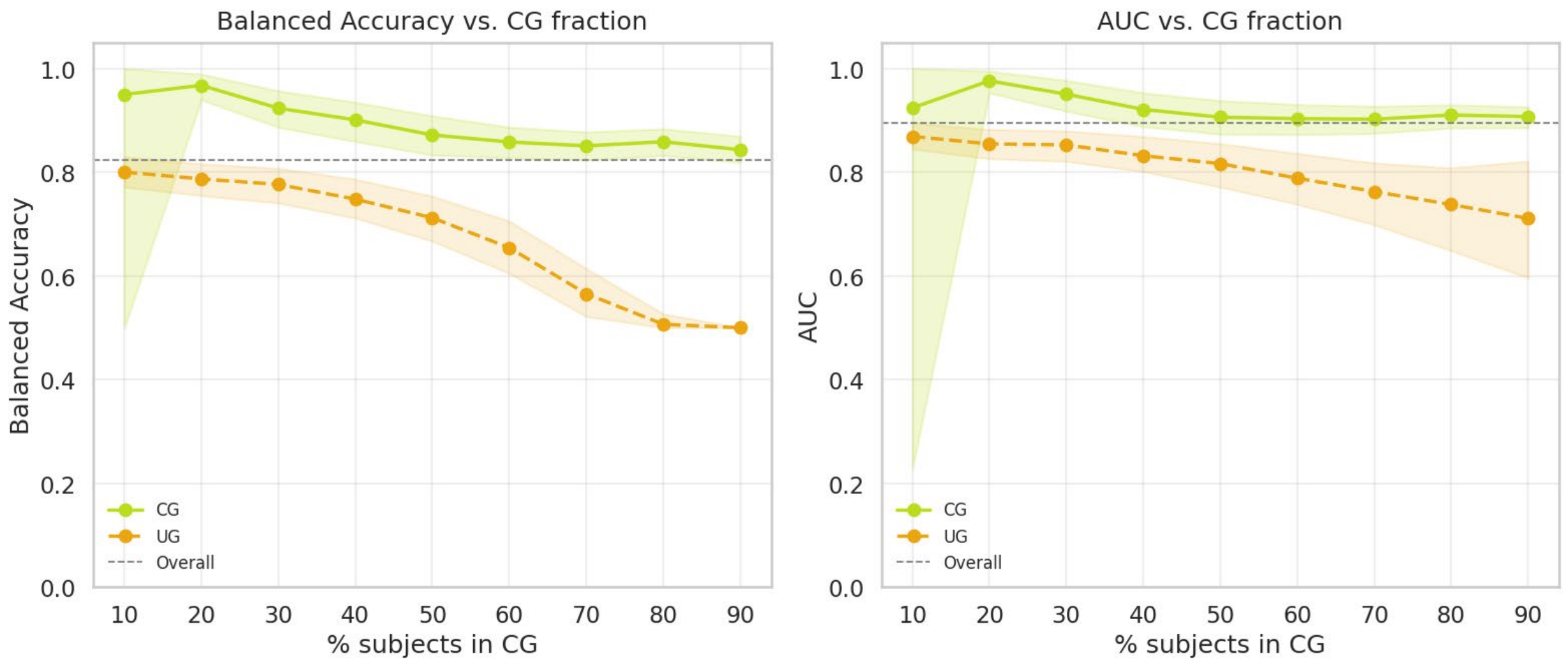

*Figure 4. Performance analysis based on uncertainty. Balanced accuracy (left) and Area under the curve (AUC, right), are plotted as a function of the percentage of subjects retained in the certain group (CG) and uncertain group (UG). Solid lines indicate CG performance, dashed lines indicate UG performance, and the horizontal dashed line represents overall model performance. Shaded regions denote 95% bootstrap confidence intervals.*

## 3.6. Decision curve analysis:

We further assessed the clinical utility of the classification model using decision curve analysis (DCA) on the combined EuCanImage + TCIA_CRLM test set (Figure 5). The model exhibited a consistently higher net benefit than both 'treat-all' and 'treat-none' strategies across a broad range of threshold probabilities (~0.30–0.75). The optimum net benefit was observed around a threshold of about 0.5, which implies that the model provides the greatest clinical advantage when the default probability threshold is considered. At low thresholds (< 0.3), the 'treat-all' and model curves are overlapping showing that the model has no advantage in that range. However, the model never fell under the negative net benefit of a 'treat-all' strategy at higher thresholds, which showed that it could help reduce unnecessary interventions while maintaining sensitivity for CRLM detection. The net benefit of the model became close to zero at very high thresholds (> 0.8), implying that there is little value at very high thresholds unless highly confident predictions are being used to make decisions.

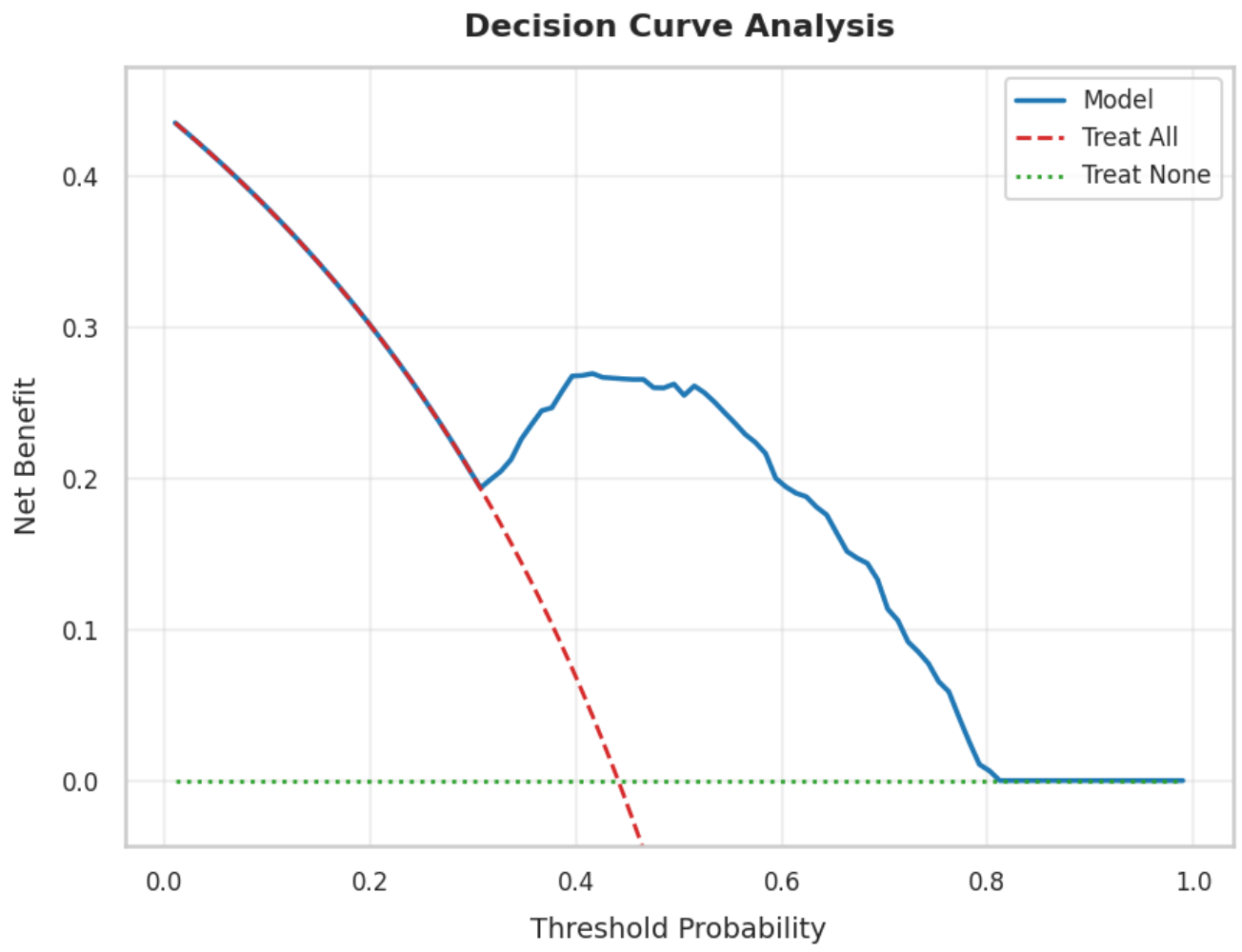

*Figure 5. Decision curve analysis (DCA) on the combined EuCanImage + TCIA_CRLM test set. Net benefit is plotted against threshold probability for the classification model (blue), compared with "treat-all" (red dashed) and "treat-none" (green dotted) strategies. The model shows superior net benefit across thresholds between approximately 0.30 and 0.75, peaking near 0.5.*

## 3.7. Grad-CAM explainability analysis:

To assess model explainability, Grad-CAM heatmaps were generated for scan-level classification predictions (see Figure 6). Every panel has a corresponding CT slice at its top and the Grad-CAM visualization at its bottom, with warmer colors representing the regions that have the highest contribution to the label being predicted. The examples show model attention in cases of CRLM. The highlighted regions in panels A and B have directly overlapping visible hepatic lesions and are associated with high-confidence CRLM predictions (0.8). In comparison, panels C and D depict weaker patterns of attention. In panel C, the heatmap has a diffuse distribution throughout the abdomen, with the strongest activation observed outside the liver, even though the ground-truth label is negative, which corresponds with the lower predicted probability (0.6). In panel D, the attention is localized, although it covers the lesion, there are also some false positive regions, which implies a lower-confidence positive prediction (0.4). These results show that the model tends to focus on clinically relevant hepatic regions in true positive cases, where lower-confidence predictions are linked with more diffuse or off-target activation.

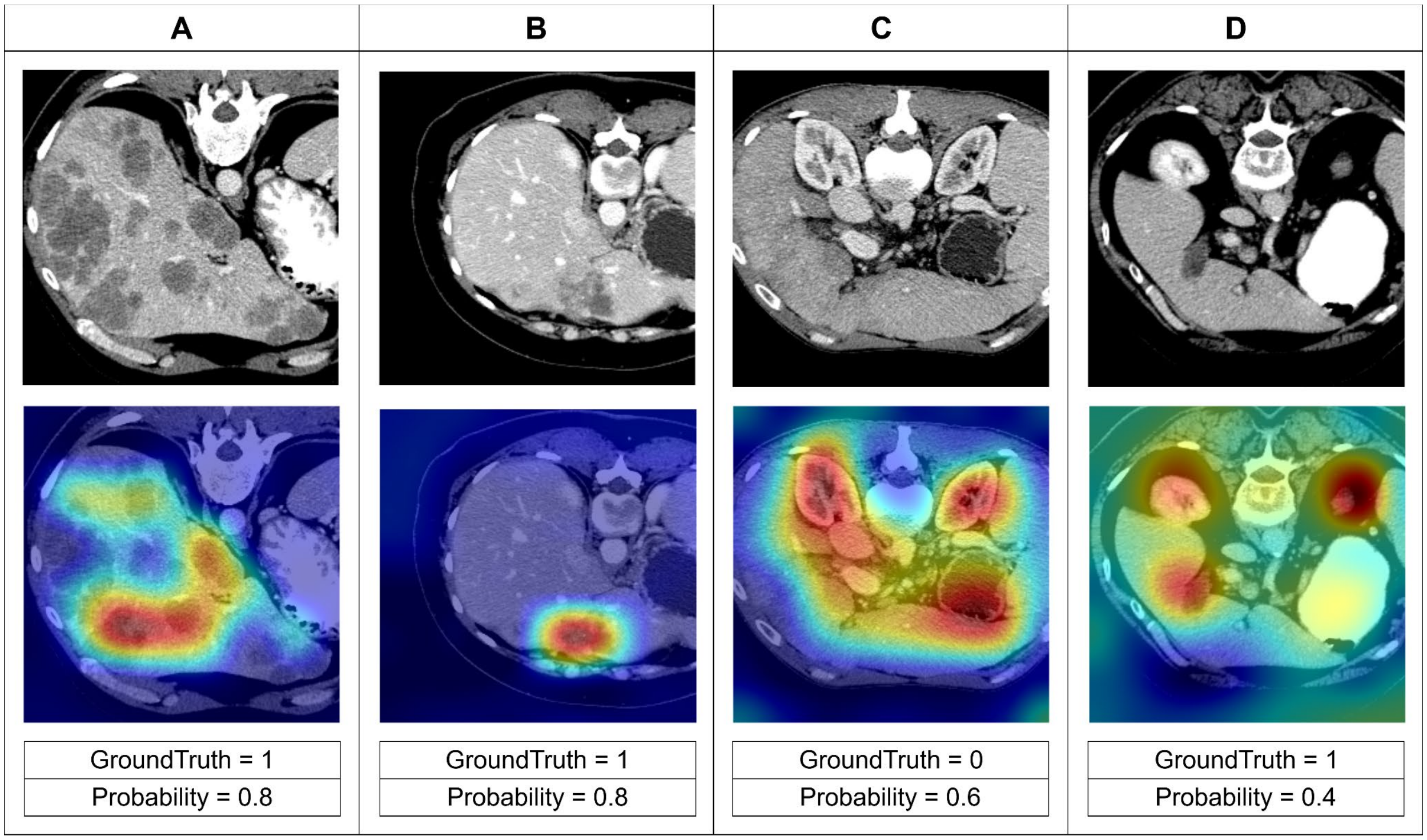

*Figure 6. Grad-CAM visualizations for scan-level classification predictions. Each panel shows the original CT slice (top) and the corresponding Grad-CAM heatmap (bottom), with ground-truth labels and predicted probabilities displayed below. Panels A and B demonstrate alignment between model attention and visible liver metastases, while panels C and D illustrate less precise activation, including off-target focus on non-hepatic structures (C) or partial lesion coverage (D).*

## 3.8. Lesion-Level Detection Performance:

The quantitative analysis for lesion detection was performed on the TCIA-CRLM dataset (Figure 7), which contains detailed lesion annotations. In the TCIA_CRLM test set, the model was able to detect 331 of 479 ground-truth lesions (69.1%) at an IoU threshold of 0.025 and with a minimum lesion size of 50 voxels (≈ 0.125 $cm^3$). Detection performance increased consistently with lesion size, from 30.0% (36/120) in Q1 (≤ 0.80 $cm^3$) to 98.3% (118/120) in Q4 (> 11.49 $cm^3$). The total number of predictions made by the model was 588, with 257 false positives, a false positive per image rate of 1.3, and a false positive per lesion rate of 0.54. In the case of the EuCanImage test set, only a subset of the lesions were manually labelled, thus there was no way to report reliable quantitative metrics on the lesion level. The results of detection on this cohort were checked qualitatively and were applied mostly to validate model generalizability to other sites.

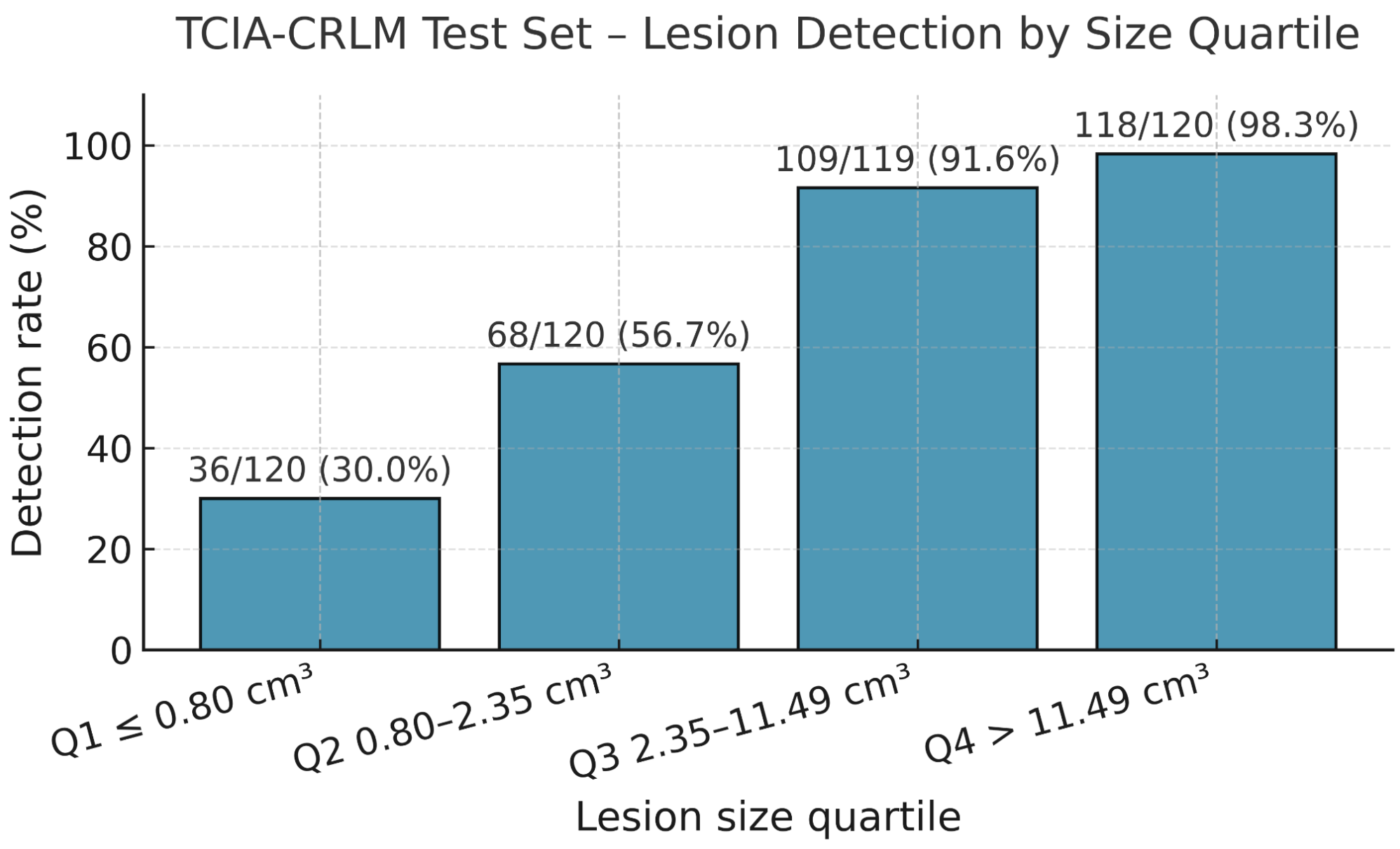

Figure 7: *Lesion detection rates stratified by lesion size quartile on the TCIA_CRLM test set. Each bar shows the number of detected lesions relative to the total ground-truth lesions within that quartile, with corresponding percentages annotated above.*

## 3.9. Cross-task consistency (TCIA):

Lesion-level detection performance improved in subsets defined by classification accuracy and confidence. In correctly classified cases, sensitivity increased from 69.1% to 71.0%, and in the high-confidence correctly classified subset (CG ∩ Correct), sensitivity reached 74.0%. False-positive rate per image dropped from 1.30 to 1.22. These results indicate that higher classification confidence corresponds to more accurate lesion localization and fewer spurious detections, demonstrating coherence between classification and detection model behavior. Refer to Table 5.

Table 5. Lesion-level detection performance on the TCIA_CRLM dataset across evaluation subsets. Detection sensitivity and false-positive rates are reported for the full test set, correctly classified cases, and the high-confidence subset (CG ∩ Correct). Detection performance improved with classification confidence, with sensitivity increasing from 69.1% to 74.0% and decreasing false positive rates.

| Metric | Overall (n = 197) | Correct (n = 167) | CG (80%) ∩ Correct (n = 134) |
|---|---|---|---|
| **Sensitivity (IoU > 0.025)** | 69.1 % | 71.0 % | **74.0 %** |
| **FPR / image** | 1.30 | 1.34 | **1.22** |
| **FPR / lesion** | 0.54 | 0.52 | **0.48** |

## 3.10. Qualitative detection analysis:

Representative examples of detection outputs from all participating EuCanImage centres and the external TCIA_CRLM dataset are presented in Figure 8. Ground-truth lesion annotations are outlined in green, and model-predicted bounding boxes are shown in red. Across cohorts, the detection model consistently localized both small and large CRLM, capturing diverse appearances and contrast levels. However, performance was less reliable for subtle, low-contrast, or peripherally located lesions, some of which were missed entirely or only partially overlapped with the reference annotation. In several cases, false-positive predictions were observed near vascular or biliary structures, reflecting residual challenges in differentiating small lesions from anatomical variations. These examples illustrate the model's cross-site generalization and highlight both successful detections and typical failure modes.

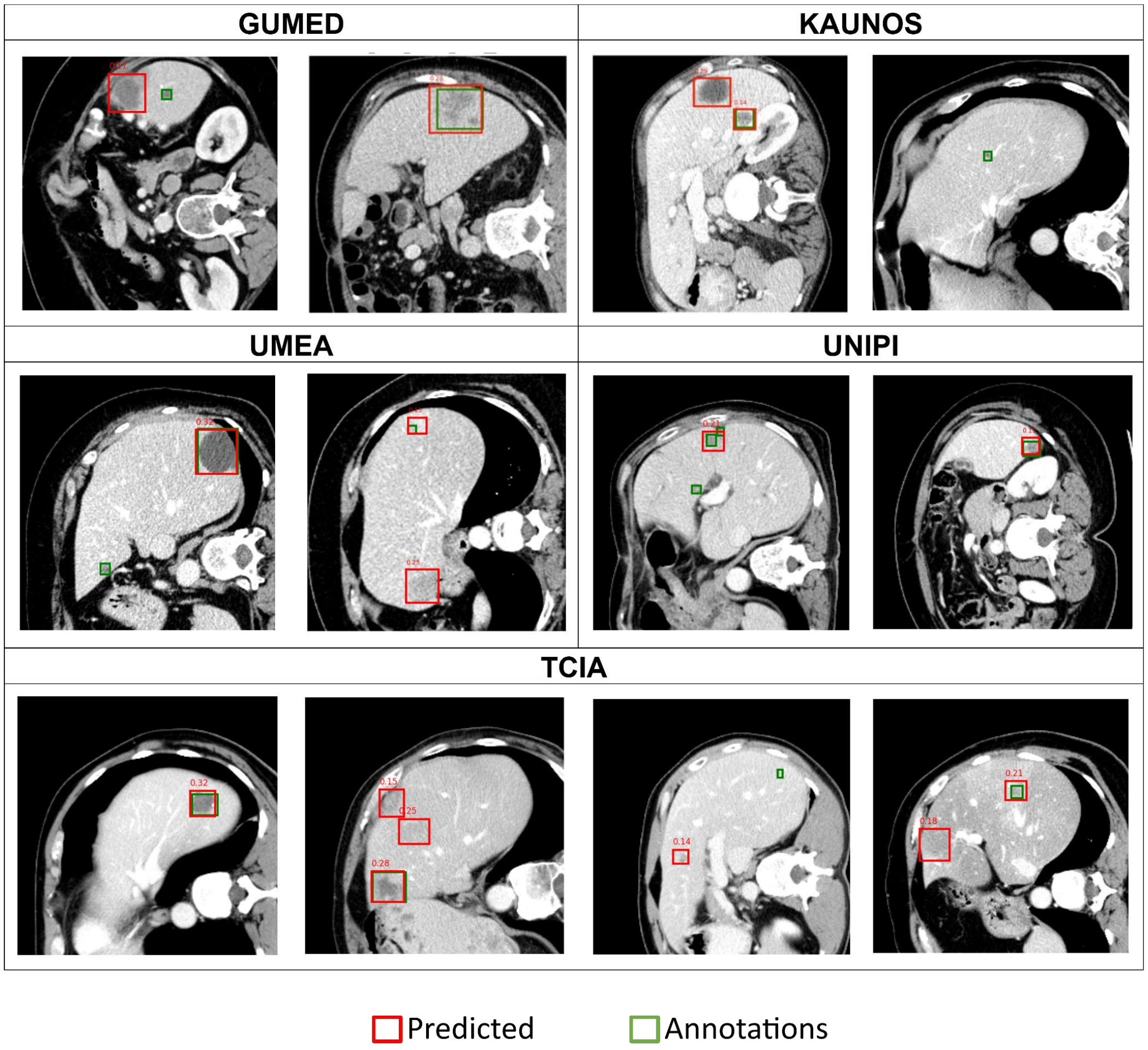

*Figure 8: Qualitative detection results across EuCanImage centres and the external TCIA_CRLM dataset. Examples from GUMED, KAUNOS, UMEA, UNIPI, and TCIA cohorts are shown. Ground-truth annotations are displayed in green, and model predictions in red. The figure demonstrates accurate detections, partial overlaps, and typical false-positive or missed cases.*

## 3.11. RQS 2.0 Assessment:

The overall methodological quality and readiness of the proposed AI pipeline were evaluated using the Radiomics Quality Score (RQS) 2.0 framework [43]. The study achieved a total of 28 out of 35 points, corresponding to a Radiomics Readiness Level (RRL) of 6, and for the deep learning track, which reflects strong methodological transparency, reproducibility, and partial clinical readiness. The scoring captured key strengths across data preparation, model development, validation, and trustworthiness. The cumulative progression of achieved versus maximum attainable scores per readiness level is shown in Figure 9, illustrating consistent methodological completeness up to RRL-6, where requirements for calibration, interpretability, and external validation were satisfied. A detailed explanation of each criterion, with supporting evidence, is provided in Supplementary Table 2.

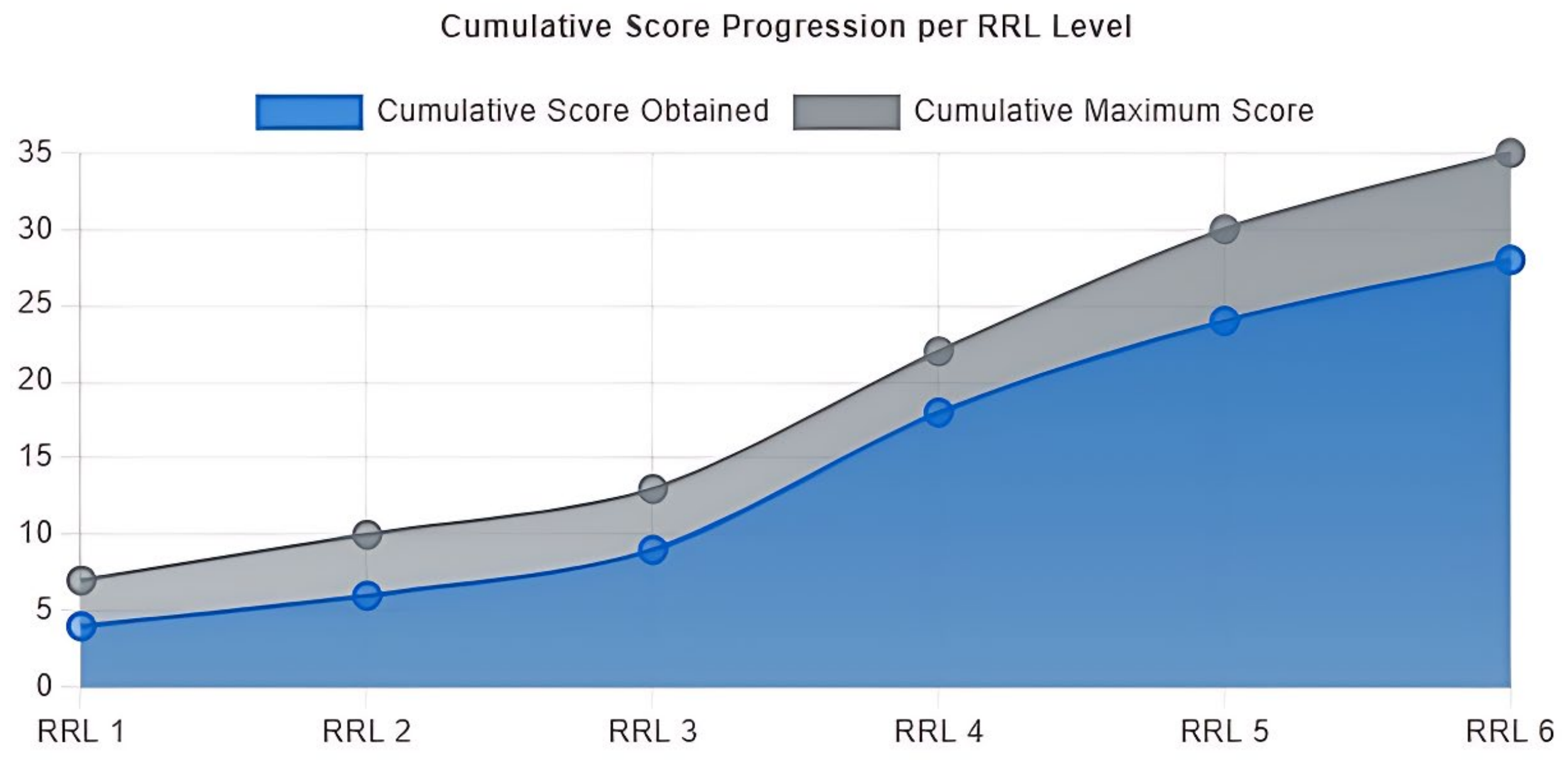

*Figure 9. Cumulative Radiomics Quality Score (RQS 2.0) progression across Radiomics Readiness Levels (RRL 1–6). The study achieved an RQS 2.0 score of 28/35, corresponding to RRL-6, indicating high methodological rigor and partial clinical readiness according to the RQS 2.0 framework.*

# 4. Discussion

In this multi-centre study, we developed and evaluated an AI pipeline for CRLM using foundation-model features for patient-level classification and lesion-level detection across heterogeneous contrast-enhanced CT. A systematic benchmark of pretrained backbones showed that UMedPT + MLP gave the most balanced performance with an AUC of 0.90 and balanced accuracy of 0.83 on the combined EuCanImage + TCIA_CRLM test cohort at an objectively selected operating point (Youden's index). The sensitivity of our classification model was also high at about 85% when applied to the external TCIACRLM data and 82% when applied to the EuCanImage hold-out test set. This highlights the clinical importance of sensitivity in the detection of CRLM because false-negative detection of true metastases may result in under-treatment or delayed treatment [44]. At the lesion level for the TCIA_CRLM dataset, the detection model achieved an overall sensitivity of 69.1% and exceeded 98% for the largest quartile. It was more challenging for small (< ~1 $cm^3$) lesions. Grad-CAM examples illustrated this pattern, with high-confidence cases showing focused hepatic attention and lower-confidence cases exhibiting diffuse or off-target activations.

Relative to previous AI research in CRLM that has largely focused on radiomics or segmentation tasks rather than integrated classification and detection, our pipeline combines foundation-model transfer, uncertainty quantification, decision-curve analysis, and explainability in a single framework. For example, Tharmaseelan et al. [45] reported that radiomics-based classifiers could discriminate colorectal vs pancreatic liver metastases with AUC ~0.87, outperforming DenseNet-121 in their task, though their focus was on primary tumor origin rather than general metastasis detection. Luo et al. [46] built radiomics signatures for disease-free survival in CRLM and demonstrated the prognostic value of radiomic features in contrast-enhanced CT, although external validation performance was modest. Montagnon et al. [47] similarly emphasized CT texture analysis in predicting chemotherapy response among CRLM lesions, finding that texture features add beyond size/volume but that performance is challenged when lesions are small. Wesdorp et al. [21] developed deep learning autosegmentation models for liver and CRLM, achieving tumor DSC ~0.86 and supporting the feasibility of automated burden assessment. Compared with these approaches, our pipeline achieved comparable or superior classification performance (AUC 0.90) in a more heterogeneous multicenter cohort, while also explicitly reporting sensitivity (>80% at optimized thresholds), which is a clinically decisive metric for CRLM management. We benchmarked several foundation models systematically and added other analyses, including uncertainty quantification, decision curve analysis, and explainability, which gives a more comprehensive view of model reliability and clinical applicability.

The key strength of this study is that it combines classification, detection, quantification of uncertainty, and explainability in one foundation model. Despite being pretrained on heterogeneous medical imaging tasks, the UMedPT backbone demonstrated strong transferability to hepatic metastasis detection, achieving robust discrimination and clinically meaningful sensitivity. Notably, the model showed consistent performance in the various test sets, which indicated that it was resilient to data heterogeneity, which is a typical problem in multi-institutional imaging. It was a large multi-center cohort of 2,437 patients across four European centres, across a wide range of scanners and protocols, and thus it is one of the most diverse datasets to have been studied in CRLM.

There were two design choices that were emphasized in our study. First, despite using all the contrast phases available to train the model to make the model more diverse and robust, the model was assessed only on images in portal-venous (PV) phases. The PV phase is widely recognized as the most informative and consistent phase for identifying liver metastases, as it provides stable parenchymal enhancement and clearer lesion delineation across scanners and institutions [48,49]. The use of all phases in the training process served as a form of data-augmentation technique, exposing the model to variations in contrast appearance and consequently improving its generalization capabilities, and testing the model on the PV phase allowed maintaining the comparability to standard clinical protocols. Second, the uncertainty-aware analysis provided further insights into model reliability. Following the framework of Alves et al. [39], we observed that cases classified as "certain" consistently yielded higher AUC and balanced accuracy compared with "uncertain" cases, validating the utility of uncertainty estimates as a triage tool. In clinical workflows, such stratification could support selective human-AI collaboration, where high-confidence cases may be safely automated while uncertain cases prompt expert review. Decision curve analysis reinforced this interpretation by showing a positive net benefit across a broad range of threshold probabilities (~0.30-0.75), with the maximum benefit observed around a threshold of 0.5. This indicates that incorporating AI-based predictions may be useful in preventing unnecessary procedures without being less sensitive to clinically relevant detection of CRLM through contrast-enhanced CT images. Together with moderate calibration (Brier 0.17; slope 3.4; intercept 0.15), these results highlight the potential of pairing model predictions with uncertainty estimates and threshold-based decision policies to enable safe and interpretable clinical decision support.

The findings of misclassifications (failure mode analysis) show that lesion size and visual subtlety remain key factors contributing to classification errors, particularly in multi-center datasets with heterogeneous acquisition protocols. These results show that UMedPT provides the most transferable and discriminative representations among the tested FMs, and that fine-tuning with an MLP head yields a practically deployable classifier with controllable operating points, high specificity on mixed cohorts, and stable behavior across centres, CT phases, and datasets, including an all-positive external subset where recall can be emphasized by threshold selection.

At the lesion level, the detection model achieved a sensitivity of around 68% across the TCIA_CRLM set. Detection performance was strongly dependent on lesion size, exceeding 90% for the largest quartile of lesions (>2-11 $cm^3$) but dropping to 30% for the smallest quartile (<1 $cm^3$). This pattern aligns with known clinical challenges in identifying small or low-contrast metastases on CT, where even expert readers exhibit reduced sensitivity [50]. Quantitative false-positive rates were assessed only on the TCIA-CRLM dataset, which includes complete lesion annotations, yielding an average of 1.3 false positives per image and 0.54 per lesion. For the EuCanImage cohort, lesion annotations were incomplete, so false positives could not be reliably estimated; however, qualitative inspection indicated similar trends, with most spurious detections occurring near vascular or biliary structures rather than within healthy parenchyma.

Cross-task analysis further revealed that detection precision improved in high-confidence correctly classified cases. Within this subset, lesion sensitivity increased from 69.1% to 74.0%, while the false-positive rate decreased from 1.30 to 1.22 per image, demonstrating that higher classification confidence corresponded to more accurate lesion localization and fewer spurious detections. This coherence between classification confidence and detection accuracy underscores that uncertainty information is not only a proxy for case-level reliability but also reflects underlying feature consistency across tasks.

Grad-CAM explainability also improved interpretability, showing a high correlation between the heatmaps and annotated lesions in high-confidence cases. In contrast, diffuse or off-target activations were often associated with uncertain predictions, reinforcing the complementarity of uncertainty estimation and visual explainability. These results are in line with a clinically oriented paradigm where foundation models not only give good diagnostic performance but also give calibrated confidence and interpretable outputs that are relied on in decision support..

However, there are a few limitations to this study. Sensitivity for small lesions (first quartile) remains modest. Small lesions were a principal failure mode in detection, reflecting known CT constraints for sub-centimeter disease; false positives clustered around vascular/biliary structures. Second, EuCanImage lesion annotations are partial, precluding reliable lesion-wise sensitivity on that cohort; TCIA_CRLM therefore served as the quantitative reference for detection. While this does not diminish the model's high specificity at the patient level, it highlights the need for refined post-processing or uncertainty-guided filtering to suppress spurious predictions. Third, in Grad-CAM visualization, off-target attention in lower-confidence cases suggests that the model sometimes leverages non-hepatic cues or background context, which could mislead in certain cases. Finally, although our data encompassed multiple European centres, broader external validation beyond the EuCanImage and TCIA_CRLM cohorts, ideally through prospective evaluation, will be critical to confirm generalizability and clinical applicability before deployment.

Future directions should focus on improving the detection of small lesions, which remain the principal challenge in both clinical and automated CRLM assessment. Potential strategies include higher-resolution or multi-phase contrast CT inputs, specialized small-lesion-aware detection heads, and post-processing guided by uncertainty estimates to suppress spurious predictions. Additionally, cases could be explored where patients with cardiac dysfunction may show heterogeneous enhancement of the liver during PV phase, potentially increasing the likelihood of false negative predictions (especially in failure mode analysis). This effect has been indirectly observed in previous studies [51,52] where early or delayed PVP timing and perfusion variability can significantly reduce lesion detection in CRLM imaging. Integration of clinical and pathological metadata (e.g., tumour markers, genetic profiles, treatment history) with imaging features could further enhance patient-level classification and prognostic modeling. For interpretability, extending beyond Grad-CAM to attention-based or perturbation-driven or spatial-variance-driven explainers may provide higher-fidelity insight into model reasoning. Finally, prospective, multi-centre validation and clinical trials will be essential to confirm performance in real-world diagnostic workflows and to quantify the clinical utility of uncertainty-informed AI support. Overall, these directions align with a broader goal of building trustworthy, interpretable, and clinically actionable foundation-model pipelines for CRLM detection and staging on CT.

# 5. Conclusion

In conclusion, this study demonstrates that foundation model-based pipelines, when systematically benchmarked and fine-tuned with appropriate preprocessing and interpretability strategies, can deliver robust and generalizable performance for both patient-level classification and lesion-level detection of colorectal liver metastases in multi-center CT data. The integration of classification, detection, calibration, uncertainty, and explainability within a single unified framework represents a significant advance toward trustworthy AI in oncology imaging. Despite persistent challenges, most notably the limited sensitivity for small or subtle lesions and the occurrence of false positives near vascular structures, the proposed approach achieves clinically relevant recall and interpretability across heterogeneous datasets. The inclusion of decision-curve analysis and uncertainty-aware evaluation provides a transparent assessment of clinical benefit and model reliability, addressing key aspects of translational readiness. Collectively, these results highlight the feasibility of deploying foundation model-driven methods as a critical step toward clinically applicable, transparent, and reliable AI tools for CRLM detection, staging, and treatment planning in contrast-enhanced CT.

# Data and code availability

Data access information is provided via the EuCanImage page (https://eucanimage.eu/access-eucanimage-data-portal-today/). The codes and data analysis scripts are available on Github repository https://github.com/shruti26mali/Multicentre-CRLM-Classification

# Grants and funding

Authors acknowledge financial support from ERC advanced grant (ERC-ADG-2015 n° 694812 - Hypoximmuno), ERC-2020-PoC: 957565-AUTO.DISTINCT. Authors also acknowledge financial support from the European Union's Horizon research and innovation programme under grant agreement: EuCanImage n° 952103 (main contributor), ImmunoSABR n° 733008, CHAIMELEON n° 952172, TRANSCAN Joint Transnational Call 2016 (JTC2016 CLEARLY n° UM 2017-8295), IMI-OPTIMA n° 101034347, AIDAVA (HORIZON-HLTH-2021-TOOL-06) n°101057062, REALM (HORIZON-HLTH-2022-TOOL-11) n° 101095435, RADIOVAL (HORIZON-HLTH-2021-DISEASE-04-04) n°101057699 and EUCAIM (DIGITAL-2022-CLOUD-AI-02) n°101100633.

# Disclosures:

Disclosures from the last 36 months within and outside the submitted work: none related to the current manuscript; outside of current manuscript: grants/sponsored research agreements from Radiomics SA, Convert Pharmaceuticals and LivingMed Biotech. He received a presenter fee (in cash or in kind) and/or reimbursement of travel costs/consultancy fee (in cash or in kind) from Radiomics SA, BHV & Roche. PL has shares in the companies Radiomics SA, Convert pharmaceuticals, Comunicare, LivingMed Biotech, BHV and Bactam. PL is co-inventor of two issued patents with royalties on radiomics (PCT/NL2014/050248 and PCT/NL2014/050728), licensed to Radiomics SA; one issued patent on mtDNA (PCT/EP2014/059089), licensed to ptTheragnostic/DNAmito; one non-issued patent on LSRT (PCT/ P126537PC00, US: 17802766), licensed to Varian; three non-patented inventions (softwares) licensed to ptTheragnostic/DNAmito, Radiomics SA and Health Innovation Ventures and two non-issued, non-licensed patents on Deep Learning-Radiomics (N2024482, N2024889). He confirms that none of the above entities were involved in the preparation of this paper.

# Author contributions

conceptualization, S.A.M. and Z.S.; methodology, S.A.M. and Z.S.; software, S.A.M., and Z.S.; validation, S.A.M., and Z.S.; formal analysis, S.A.M., and Z.S.; investigation, S.A.M., and Z.S.; writing - original draft preparation, S.A.M., and Z.S.; writing - review and editing, S.A.M., Z.S., Y.Z., A.A., X.Z., H.C.W., M.B., K.R., J.K., L.F., R.F., R.L.M., and P.L.; supervision, Z.S., H.C.W., and P.L.; funding acquisition, P.L.

All authors have read and agreed to the published version of the manuscript.

# Supplementary material

## Effect of ComBat Harmonization on UMedPT Deep Features

To evaluate whether scanner- or site-specific effects influenced the deep features extracted from the UMedPT foundation model, feature-level harmonization was performed using the ComBat method [53,54]. Harmonization was applied across centres (KAU, GUMED, UMEA, UNIPI) using the open-source neuroCombat implementation, with parameters estimated from the training set and transferred to validation and test data to prevent data leakage.

Following harmonization, the logistic regression model was retrained using the ComBat-adjusted features. On the EuCanImage test set, the model achieved a ROC AUC of 0.72 (95% CI: 0.67–0.76) and a balanced accuracy of 0.67 (95% CI: 0.63–0.70), which was lower than the baseline LR model trained on unharmonized UMedPT deep features (AUC = 0.81, Table 3). These results indicate that harmonization did not improve performance, suggesting that the pretrained foundation model had already learned domain-invariant representations robust to inter-site variability.

Consequently, the unharmonized UMedPT features were retained for all downstream analyses. Nonetheless, performing this comparison demonstrates adherence to reproducibility and harmonization principles recommended in the Radiomics Quality Score (RQS) 2.0 framework [43].

***Supplementary Table 1.*** *Baseline clinical characteristics of the EuCanImage CRLM cohort, stratified by the contributing centre. Values are reported as n (%) unless otherwise stated. These variables, although not directly analyzed in the main text, provide context on comorbidities and liver pathologies within the cohort.*

| | KAUNOS | GUMED | UMEA | UNIPI |
|---|---|---|---|---|
| *Clinical variables* | | | | |
| **Diagnosis** | | | | |
| Histopathology | 59 (12.0%) | 75 (8.8%) | - | - |
| Imaging | 431 (88.0%) | 773 (91.2%) | - | 698 (100.0%) |
| **Comorbidity** | | | | |
| Yes | 141 (28.8%) | 196 (23.1%) | - | 7 (1.0%) |
| No | 338 (69.0%) | 286 (33.7%) | - | 6 (0.9%) |
| unknown | 11 (2.2%) | 366 (43.2%) | - | 685 (98.1%) |
| **Biliary cystadenoma** | | | | |
| Yes | - | - | - | 1 (0.1%) |
| No | 191 (39.0%) | 286 (33.7%) | - | 41 (5.9%) |
| unknown | 299 (61.0%) | 562 (66.3%) | - | 638 (91.4%) |
| **Biloma** | | | | |
| Yes | - | 2 (0.2%) | - | - |
| No | 193 (39.4%) | 286 (33.7%) | - | 42 (6.0%) |
| unknown | 297 (60.6%) | 560 (66.0%) | - | 656 (94.0%) |
| **Cholangiocarcinomas (bile duct cancers)** | | | | |
| Yes | 3 (0.6%) | 2 (0.2%) | - | - |
| No | 190 (38.8%) | 286 (33.7%) | - | 42 (6.0%) |
| unknown | 297 (60.6%) | 560 (66.0%) | - | 656 (94.0%) |
| **Dysplastic liver nodule** | | | | |
| Yes | 7 (1.4%) | - | - | - |
| No | 185 (37.8%) | 286 (33.7%) | - | 42 (6.0%) |
| unknown | 298 (60.8%) | 562 (66.3%) | - | 656 (94.0%) |
| **Fibrolamellar hepatocellular carcinoma** | | | | |
| Yes | - | - | - | - |
| No | 192 (39.2%) | 286 (33.7%) | - | 42 (6.0%) |
| unknown | 298 (60.8%) | 562 (66.3%) | - | 656 (94.0%) |
| **Focal hepatic steatosis** | | | | |
| Yes | 10 (2.0%) | 31 (3.7%) | - | - |
| No | 180 (36.7%) | 286 (33.7%) | - | 42 (6.0%) |
| unknown | 300 (61.2%) | 531 (62.6%) | - | 656 (94.0%) |
| **Focal nodular hyperplasia** | | | | |
| Yes | 1 (0.2%) | 2 (0.2%) | - | - |
| No | 192 (39.2%) | 286 (33.7%) | - | 42 (6.0%) |
| unknown | 297 (60.6%) | 560 (66.0%) | - | 656 (94.0%) |
| **Hemangioma** | | | | |
| Yes | 44 (9.0%) | 38 (4.5%) | - | 3 (0.4%) |
| No | 149 (30.4%) | 286 (33.7%) | - | 41 (5.9%) |
| unknown | 297 (60.6%) | 524 (61.8%) | - | 656 (94.0%) |
| **Hepatic abscess** | | | | |
| Yes | 2 (0.4%) | 2 (0.2%) | - | - |
| No | 191 (39.0%) | 286 (33.7%) | - | 42 (6.0%) |
| unknown | 297 (60.6%) | 560 (66.0%) | - | 656 (94.0%) |
| **Hepatic adenoma** | | | | |
| Yes | 2 (0.4%) | - | - | - |
| No | 191 (39.0%) | 286 (33.7%) | - | 42 (6.0%) |
| unknown | 297 (60.6%) | 562 (66.3%) | - | 656 (94.0%) |
| **Hepatic angiosarcoma** | | | | |
| Yes | - | - | - | - |
| No | 193 (39.4%) | 286 (33.7%) | - | 42 (6.0%) |
| unknown | 297 (60.6%) | 562 (66.3%) | - | 656 (94.0%) |
| **Hepatic hydatid cyst** | | | | |
| Yes | 1 (0.2%) | - | - | - |
| No | 192 (39.2%) | 286 (33.7%) | - | 42 (6.0%) |
| unknown | 297 (60.6%) | 562 (66.3%) | - | 656 (94.0%) |
| **Hepatocellular carcinoma** | | | | |
| Yes | 3 (0.6%) | 1 (0.1%) | - | - |
| No | 190 (38.8%) | 286 (33.7%) | - | 20 (2.9%) |
| unknown | 297 (60.6%) | 561 (66.2%) | - | 678 (97.2%) |
| **Liver cirrhosis** | | | | |
| Yes | 27 (5.5%) | 3 (0.4%) | - | 1 (0.1%) |
| No | 166 (33.9%) | 286 (33.7%) | - | 41 (5.9%) |
| unknown | 297 (60.6%) | 559 (65.9%) | - | 656 (94.0%) |
| **Primary hepatic lymphoma** | | | | |
| Yes | - | - | - | - |
| No | 193 (39.4%) | 286 (33.7%) | - | 42 (6.0%) |
| unknown | 297 (60.6%) | 562 (66.3%) | - | 656 (94.0%) |
| **Simple hepatic cysts** | | | | |
| Yes | 86 (17.6%) | 183 (21.6%) | - | 5 (0.7%) |
| No | 104 (21.2%) | 286 (33.7%) | - | 38 (5.4%) |
| unknown | 300 (61.2%) | 379 (44.7%) | - | 655 (93.9%) |
| **Transient hepatic attenuation differences** | | | | |
| Yes | - | - | - | - |
| No | 191 (39.0%) | 286 (33.7%) | - | 42 (6.0%) |
| unknown | 299 (61.0%) | 562 (66.3%) | - | 656 (94.0%) |
| **Pathological T staging (primary tumor)** | | | | |
| 0 (pTX) | 1 (0.2%) | - | - | |
| 1 (pT0) | - | 1 (0.1%) | - | |
| 3 (pT1) | 1 (0.2%) | 3 (0.4%) | - | 6 (0.9%) |
| 4 (pT2) | 8 (1.6%) | 16 (1.9%) | - | 7 (1.0%) |
| 5 (pT3) | 59 (12.0%) | 107 (12.6%) | - | 19 (2.7%) |
| 6 (pT4) | 9 (1.8%) | 6 (0.7%) | - | 4 (0.6%) |
| 7 (pT4a) | 17 (3.5%) | 29 (3.4%) | - | 5 (0.7%) |
| 8 (pT4b) | 3 (0.6%) | 9 (1.1%) | - | 3 (0.4%) |
| unknown | 392 (80.0%) | 677 (79.8%) | - | 654 (93.7%) |
| **Clinical T staging (primary tumor)** | | | | |
| 0 (cTX) | 31 (6.3%) | 90 (10.6%) | 3 (0.7%) | 1 (0.1%) |
| 3 (cT1) | 1 (0.2%) | 2 (0.2%) | - | - |
| 4 (cT2) | 11 (2.2%) | 10 (1.2%) | 84 (20.9%) | - |
| 5 (cT3) | 60 (12.2%) | 54 (6.4%) | 208 (51.9%) | 19 (2.7%) |
| 6 (cT4) | 22 (4.5%) | 20 (2.4%) | 93 (23.2%) | 8 (1.1%) |
| 7 (cT4a) | 12 (2.4%) | - | - | 2 (0.3%) |
| 8 (cT4b) | 4 (0.8%) | 6 (0.7%) | - | 4 (0.6%) |
| unknown | 349 (71.2%) | 666 (78.5%) | 13 (3.2%) | 664 (95.1%) |
| **Pathological N staging (primary tumor)** | | | | |
| 0 (pNX) | 2 (0.4%) | 1 (0.1%) | - | - |
| 1 (pN0) | 23 (4.7%) | 46 (5.4%) | - | 7 (1.0%) |
| 2 (pN1) | 48 (9.8%) | 66 (7.8%) | - | 17 (2.4%) |
| 3 (pN2) | 25 (5.1%) | 58 (6.8%) | - | 12 (1.7%) |
| unknown | 392 (80.0%) | 677 (79.8%) | - | 662 (94.8%) |
| **Clinical N staging (primary tumor)** | | | | |
| 0 (cNX) | 35 (7.1%) | 97 (11.4%) | - | - |
| 1 (cN0) | 12 (2.4%) | 14 (1.7%) | - | 10 (1.4%) |
| 2 (cN1) | 28 (5.7%) | 36 (4.2%) | - | 14 (2.0%) |
| 3 (cN2) | 66 (13.5%) | 35 (4.1%) | - | 18 (2.6%) |
| unknown | 349 (71.2%) | 666 (78.5%) | - | 656 (94.0%) |
| **Pathological M staging (primary tumor)** | | | | |
| 0 (pM0) | 1 (0.2%) | 137 (16.2%) | - | 6 (0.9%) |
| 1 (pM1) | 63 (12.9%) | 30 (3.5%) | - | 21 (3.0%) |
| 2 (pM1a) | 5 (1.0%) | 1 (0.1%) | - | 14 (2.0%) |
| 3 (pM1b) | - | 3 (0.4%) | - | 1 (0.1%) |
| 4 (pM1c) | - | - | - | 1 (0.1%) |
| unknown | 421 (85.9%) | 677 (79.8%) | - | 655 (93.8%) |
| **Clinical M staging (primary tumor)** | | | | |
| 0 (cM0) | 4 (0.8%) | 112 (13.2%) | - | 5 (0.7%) |
| 1 (cM1) | 155 (31.6%) | 46 (5.4%) | - | 16 (2.3%) |
| 2 (cM1a) | 3 (0.6%) | 4 (0.5%) | - | 2 (0.3%) |
| 3 (cM1b) | 5 (1.0%) | 7 (0.8%) | - | 5 (0.7%) |
| 4 (cM1c) | 1 (0.2%) | 13 (1.5%) | - | 3 (0.4%) |
| unknown | 322 (65.7%) | 666 (78.5%) | - | 667 (95.6%) |

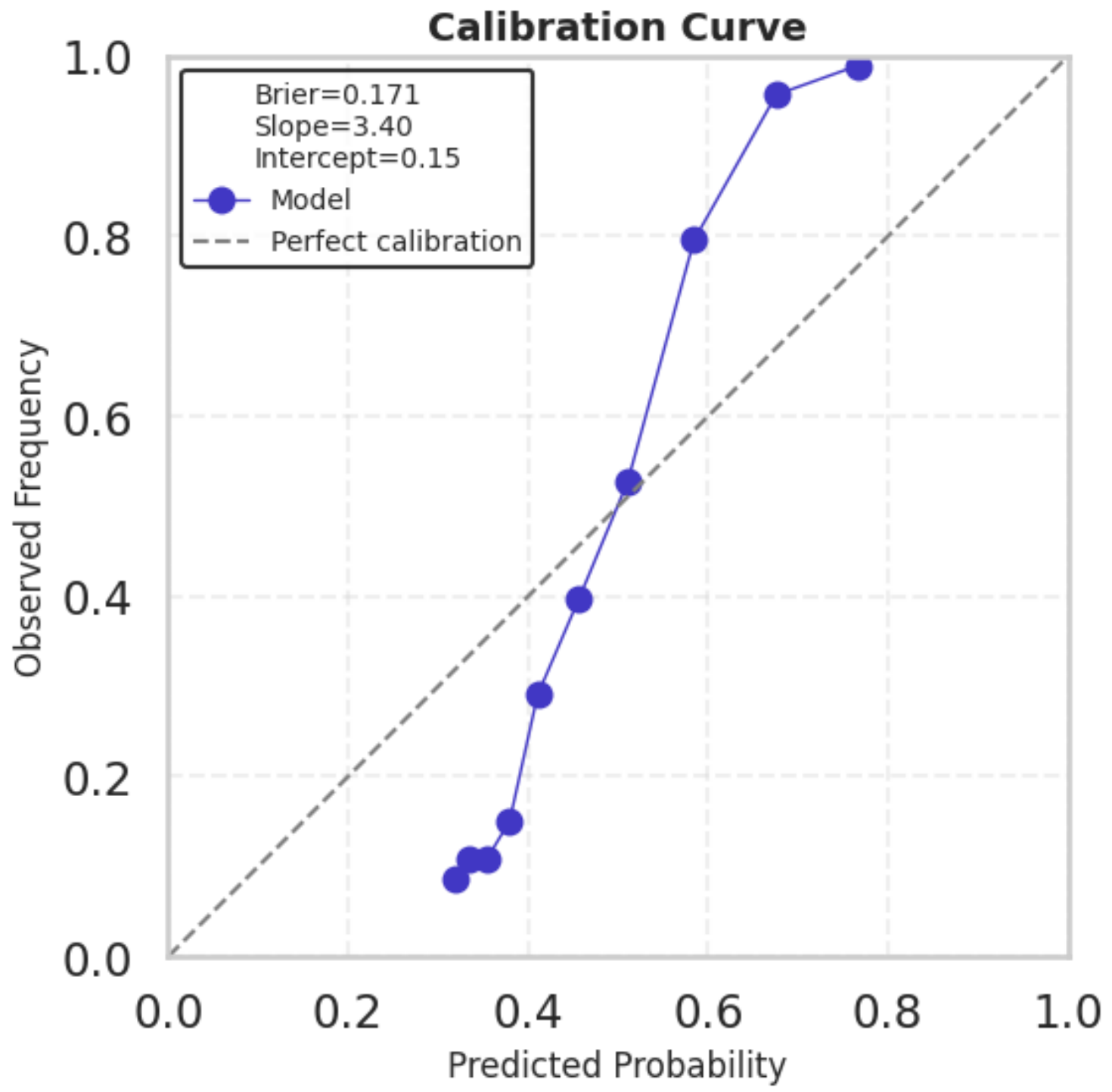

***Figure 1.*** *Calibration curve of the best-performing model on the combined test set. The diagonal gray dashed line indicates perfect calibration. The solid line shows observed versus predicted probabilities averaged across the prospective validation folds. The model achieved a Brier score of 0.171, calibration slope = 3.40, and intercept = 0.15.*

***Supplementary Table 2.*** *Detailed Radiomics Quality Score (RQS 2.0) evaluation for the proposed multi-center CRLM classification–detection framework. The study achieved an RQS 2.0 score of 28/35 (Radiomics Readiness Level 6), reflecting strong methodological transparency, validation, and interpretability consistent with the RQS 2.0 framework*

| No. | Criteria | Selected Option | Points | Explanation |
|---|---|---|---|---|
| | **RRL 1 - Foundational Exploration** | | | |
| 1 | Unmet Clinical Need – Unmet clinical need (UCN) defined.<br>● UCN is agreed upon and defined by more than one center.<br>● UCN is defined using an established consensus method such as the Delphi method. | Implemented: Delphi method (+2) | 2 | UCN defined and endorsed via consensus across 4 EuCanImage centres, reflecting multi-centre agreement on UCN in CRLM classification and detection |
| 2 | Hardware Description – Detailed description of the imaging hardware used, including model, manufacturer, and technical specifications. | Not implemented | 0 | Detailed scanner/hardware metadata unavailable for the participating centres |
| 3 | Image Protocol Quality – Five levels of image protocol quality for TRIAC:<br>● Level 0: Protocol not formally approved.<br>● Level 1: Approved with a reference number in the institutional archive.<br>● Level 2: Approved with formal quality assurance (recommended minimum for prospective trials). | Not implemented | 0 | No formal or standardized imaging protocol across centres; institutional approval documentation not available |
| | ● Level 3: Established internationally; published in guidelines and peer-reviewed papers.<br>● Level 4: Future proof (follows TRIAC Level 3, FAIR principles, retains raw data). | | | |
| 4 | Inclusion and Exclusion Criteria – Detailed criteria for patient selection in studies, including rationale. | Implemented (+1) | 1 | Detailed inclusion/exclusion criteria were defined and are documented in the Methods (Dataset) section |
| 5 | Diversity and Distribution – Identify potential biases before the project (demographics, socioeconomic, geographic, medical profiles). | Implemented (+1) | 1 | Patient demographic and acquisition heterogeneity were assessed (see Table 2) |
| **RRL 2 - Data Preparation** | | | | |
| 7 | Preprocessing of Images – Apply steps to standardize images with clear reasoning. | Implemented (+1) | 1 | Preprocessing steps (resampling, normalization, cropping) are fully described in the Methods section |
| 8 | Harmonization – Use image-level (e.g. CycleGANs) or feature-level (e.g. ComBat) harmonization techniques. | Implemented (+1) | 1 | Feature-level harmonization via ComBat applied across sites to mitigate scanner/site bias |
| 10 | Automatic Segmentation – Use an automated segmentation algorithm for ROI definition. | Implemented (+1) | 1 | Automated segmentation was performed using TotalSegmentator for liver ROI localization (Methods) |
| **RRL 3 - Prototype Model Development** | | | | |
| 13 | HCR + DL Combination – Compare and explore the synergistic combination of handcrafted radiomics and deep learning models. | Implemented (+1) | 1 | Combination of handcrafted radiomics (HCR) and deep-learning (DL) features was not feasible due to partial annotations |
| 14 | Multivariable Analysis – Incorporate non-radiomics features (clinical, genomic, proteomic) to yield a holistic model. | Implemented (+2) | 2 | Non-radiomics (clinical) features (age, gender) were integrated with UMedPT deep features in a logistic regression model |
| **RRL 4 - Internal Validation** | | | | |
| 15 | Single Center Validation – Validation performed on data from the same institute without retraining or adapting the cut-off value. | Implemented (+1) | 1 | Internal site-wise validation was conducted to confirm within-center generalization |
| 16 | Cut-off Analyses – Identify optimal thresholds (e.g., using Youden's Index) for classification or survival analysis. | Implemented (+1) | 1 | Optimal classification threshold(s) were selected using Youden's Index and validated on held-out data |
| 17 | Discrimination Statistics – Report discrimination metrics (e.g., ROC curve, sensitivity, specificity) with significance (p-values, CIs).<br>● Statistic reported<br>● With Resampling method | Resampling method applied (+2) | 2 | Discrimination metrics (ROC, AUC, sensitivity, specificity, 95% CI) were reported, and statistical significance assessed via resampling (e.g. bootstrapping) |
| 18 | Calibration Statistics – Report calibration metrics (e.g., calibration-in-the-large, slope, plots). | Implemented (+1) | 1 | Calibration performance reported via calibration plots and calibration-in-the-large or slope metrics |
| 19 | Failure Mode Analysis – Document model limitations with examples of edge cases. | Implemented (+1) | 1 | Misclassified/edge cases were qualitatively inspected for both internal (EuCanImage) and external (TCIA_CRLM) cohorts to understand failure modes |
| 20 | Open Science and Data – Make code and data publicly available.<br>● Open scans (+1)<br>● Open segmentations (+1)<br>● Open code (+1) | All aspects (+3) | 3 | Scans, segmentations, and code are made publicly available via the EuCanImage platform and GitHub repository |
| **RRL 5 - Capability Testing** | | | | |
| 21 | Multi-center Validation – Validation with data from multiple institutes ensuring no overlap:<br>● One external institute<br>● Two or more external institutes<br>● Third-party platform with completely unseen data | One Institute (+1) | 1 | External validation was limited to the TCIA dataset; no other independent sites were used |
| 22 | Comparison with 'Current Clinical Standard' – Assess model agreement or superiority versus the current gold standard (e.g., TNM staging). | Implemented (+2) | 2 | Model performance was evaluated against expert radiologic reference labels that serve as the clinical ground truth for CRLM diagnosis, as TNM or pathological staging data were missing across most centres. |
| 23 | Comparison to Previous Work – Compare performance with published HCR signatures or DL algorithms. | Implemented (+1) | 1 | Comparison with existing literature (handcrafted radiomics ) was limited due to incompatible annotation standards and dataset overlap |
| 24 | Potential Clinical Utility – Report on the current and potential clinical application (e.g., decision curve analysis). | Implemented (+2) | 2 | Decision curve analysis (DCA) is reported to illustrate potential clinical usefulness |
| **RRL 6 - Trustworthiness Assessment** | | | | |
| 25 | Explainability – Apply explainability tools (e.g., SHAP for HCR, GradCAM for DL) to clarify model predictions. | Implemented (+1) | 1 | Explainability via Grad-CAM visualizations is reported, highlighting regions contributing to classification decisions |
| 26 | Explainability Evaluation – Conduct qualitative and quantitative evaluations of interpretability methods (e.g., checking consistency to adversarial perturbations). | Not implemented | 0 | Not implemented |
| 27 | Biological Correlates – Detect and discuss biological correlates to deepen understanding of radiomics and underlying biology. | Not implemented | 0 | Biological correlates (e.g. linking imaging features to molecular markers) were not available in the current dataset but is planned for future work |
| 28 | Fairness Evaluation and Mitigation – Evaluate model performance for biases and apply bias correction if needed.<br>● Fairness evaluated<br>● Bias correction applied | With Bias correction (+2) | 2 | Fairness analyses were performed (e.g. stratification by lesion size or site), and bias mitigation was applied to correct observed imbalances during training of models (SMOTE and Focal loss) |
| | Total = 28/35 (80%) | | | |